\documentclass[10pt,twocolumn,letterpaper]{article}
\usepackage[pagenumbers]{cvpr} 

\definecolor{cvprblue}{rgb}{0.21,0.49,0.74}
\usepackage[pagebackref,breaklinks,colorlinks,allcolors=cvprblue]{hyperref}
\usepackage{multirow}
\usepackage{tcolorbox}

\definecolor{gray}{HTML}{EFEFEF}
\definecolor{LightCyan}{rgb}{0.88,1,1}
\definecolor{deepgreen}{HTML}{009900}
\def\paperID{12365} 
\def\confName{CVPR}
\def\confYear{2025}

\newcommand{\model}{VISTA\xspace}
\newcommand{\dataset}{VISTA-400K\xspace}
\newcommand{\hrbench}{HRVideoBench\xspace}

\title{\textcolor{NavyBlue}{\model}: Enhancing Long-Duration and High-Resolution Video Understanding \\ by \textcolor{NavyBlue}{VI}deo \textcolor{NavyBlue}{S}patio\textcolor{NavyBlue}{T}emporal \textcolor{NavyBlue}{A}ugmentation}

\author{
Weiming Ren$^{1,2,3}$, Huan Yang$^{3,*}$, Jie Min$^{1}$, Cong Wei$^{1,2}$, Wenhu Chen$^{1,2,}$\thanks{Corresponding authors.}\\
$^{1}$University of Waterloo, $^2$Vector Institute, $^3$01.AI\\
{\tt\small\{w2ren,wenhuchen\}@uwaterloo.ca, hyang@fastmail.com}\\
{\normalsize \url{https://tiger-ai-lab.github.io/VISTA/}}
}

\begin{document}
\maketitle
\begin{abstract}
Current large multimodal models (LMMs) face significant challenges in processing and comprehending long-duration or high-resolution videos, which is mainly due to the lack of high-quality datasets. To address this issue from a data-centric perspective, we propose \textbf{\model}, a simple yet effective \textbf{VI}deo \textbf{S}patio\textbf{T}emporal \textbf{A}ugmentation framework that synthesizes long-duration and high-resolution video instruction-following pairs from existing video-caption datasets. \model spatially and temporally combines videos to create new synthetic videos with extended durations and enhanced resolutions, and subsequently produces question-answer pairs pertaining to these newly synthesized videos. Based on this paradigm, we develop seven video augmentation methods and curate \dataset, a video instruction-following dataset aimed at enhancing long-duration and high-resolution video understanding. Finetuning various video LMMs on our data resulted in an average improvement of 3.3\% across four challenging benchmarks for long-video understanding. Furthermore, we introduce the first comprehensive high-resolution video understanding benchmark \hrbench, on which our finetuned models achieve a 6.5\% performance gain. These results highlight the effectiveness of our framework. 
\end{abstract}
    
\section{Introduction}
\label{sec:intro}
Recent advancements in large language models (LLMs) and large multimodal models (LMMs) have brought transformative changes to video understanding tasks. Traditionally, video understanding relied on training task-specific models using domain-specific datasets 
(e.g. action recognition on Kinetics \cite{carreira2017quo}, video retrieval on MSR-VTT \cite{xu2016msr}, and video captioning on YouCook2 \cite{zhou2018towards}). 
In contrast, it is now possible to process video inputs and address diverse tasks using a single video LMM \cite{lin2023video, cheng2024videollama, maaz2023video} through instruction following. However, most current (open-sourced) video LMMs are optimized for understanding and reasoning over short, low-resolution videos. \textbf{Processing long-sequence video input, such as long or high-resolution videos, remains a significant challenge for video LMMs.}

\begin{figure}[!t]
  \centering

    \begin{subfigure}[b]{1.0\linewidth}
        \centering
        \includegraphics[width=1.0\linewidth]{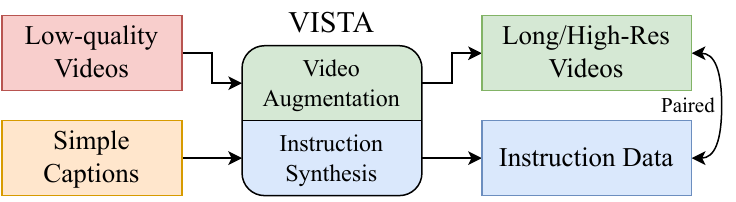}
    \end{subfigure}
    
    
    \begin{subfigure}[b]{1.0\linewidth}
        \centering
        \includegraphics[width=\textwidth]{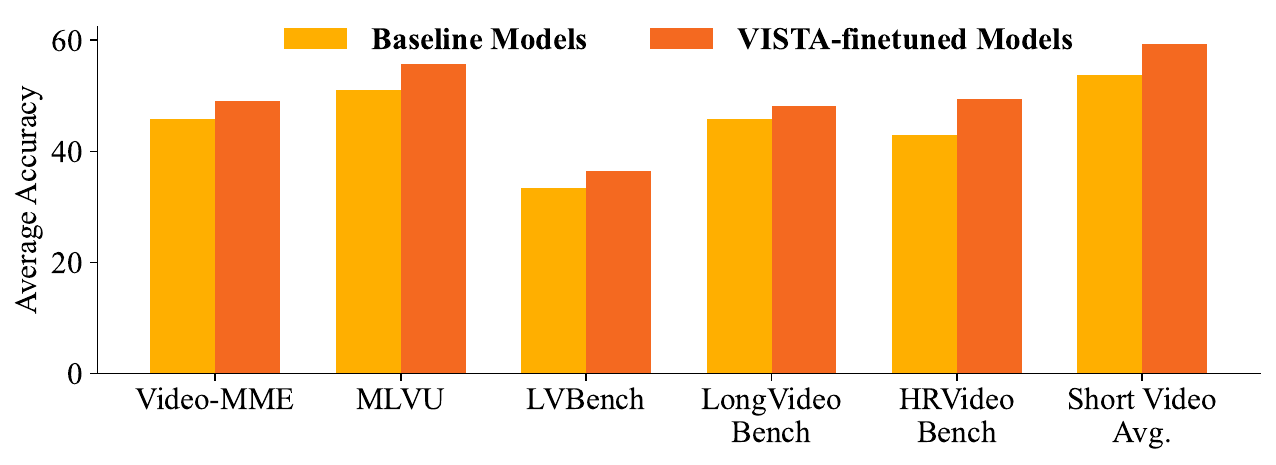}
    \end{subfigure}
  \vspace{-2em}
  \caption{\model is a simple but effective framework that generates high-quality video instruction data from existing video-caption pairs. Our \dataset dataset enhances model performances on various long and high-resolution video benchmarks.}
  \label{fig:teaser}
  \vspace{-20pt}
\end{figure}

Most efforts to enhance long-sequence video understanding have centered around designing better model architectures. Conversely, the creation of higher-quality (long/high-resolution) video instruction-following datasets remains highly under-explored. The primary challenges stem from the scarcity of publicly available (license-friendly) high-quality video-instruction data. Consequently, open-source video instruction-following datasets often face limitations such as low resolution or short duration. For example, VideoChat2 \cite{li2024mvbench} collects a video instruction-following dataset by combining videos from multiple domain-specific datasets and rewriting the instructions using ChatGPT \cite{OpenAI_ChatGPT_Website}. However, the video sources used in this dataset primarily contain short videos. ShareGPT4Video \cite{chen2024sharegpt4video} curates a video-instruction dataset containing 40K videos with detailed video captions. Nevertheless, the source videos are collected at a sparse sampling rate of 0.15 fps, and the video contents often lack motion and can appear nearly static. FineVideo \cite{Farré2024FineVideo} is a long video dataset featuring diverse video contents and high-quality metadata such as detailed captions and narrative progressions. While it offers satisfactory video durations, the dataset is limited by its resolution, as it predominantly comprises 360p videos.

Though some concurrent studies have improved LMMs' long-sequence video understanding capabilities, they often release only the model weights, hiding their video training data. For instance, Kangaroo \cite{liu2024kangaroo}'s training data contains 700K long video data, but specifics about the curation of the videos and the construction of the instructions are not provided. Similarly, Qwen2-VL \cite{wang2024qwen2}, MiniCPM-V-2.6 \cite{yao2024minicpm} and Aria \cite{li2024aria} all claim that their instruction-following dataset contains long video training data, yet without providing the sources and statistics of the video data. This lack of transparency hinders a clear understanding of what types of video instruction data truly benefit long-sequence video understanding tasks and impedes further advancements in improving existing models.

In this study, we propose \model, a simple yet effective video augmentation pipeline to synthesize \textbf{long and high-resolution video instruction data from existing video datasets}. \model leverages insights from image and video classification data augmentation techniques such as CutMix \cite{yun2019cutmix}, MixUp \cite{zhang2017mixup} and VideoMix \cite{yun2020videomix}, which demonstrate that training on synthetic data created by overlaying or mixing multiple images or videos results in more robust classifiers. Similarly, our method spatially and temporally combines videos to create (artificial) augmented video samples with longer durations and higher resolutions, followed by synthesizing instruction data based on these new videos. Our data synthesis pipeline utilizes existing public video-caption datasets, making it fully open-sourced and scalable. This allows us to construct \dataset, a high-quality video instruction-following dataset aimed at improving the long and high-resolution video understanding capabilities of video LMMs. By finetuning various video LMMs on our dataset, we observe an average of 3.3\% improvement across multiple long video understanding benchmarks. Additionally, we compile a new benchmark \hrbench that focuses on high-resolution videos. Results on \hrbench (+6.5\% after finetuning) demonstrate that our method produces models well-suited for high-resolution video understanding. Our contributions are summarized below:
\begin{itemize}
    \item [1.] We present \dataset, a high-quality synthetic video instruction-following dataset. \dataset includes challenging QA pairs designed to enhance video LMMs' ability to understand long and high-res video inputs.
    \item [2.] We collect \hrbench, a comprehensive benchmark dedicated to evaluating video LMMs' capability in understanding fine object details and subtle, localized actions within high-resolution videos.
    \item [3.] Our \model-finetuned models achieve an average gain of 3.3\% on four challenging long-video benchmarks and 6.5\% on \hrbench compared to the vanilla models. Our ablation study indicates that disabling our proposed video augmentations significantly reduces performance.
\end{itemize}

\begin{figure*}[]
  \centering
  \includegraphics[width=1.0\textwidth]{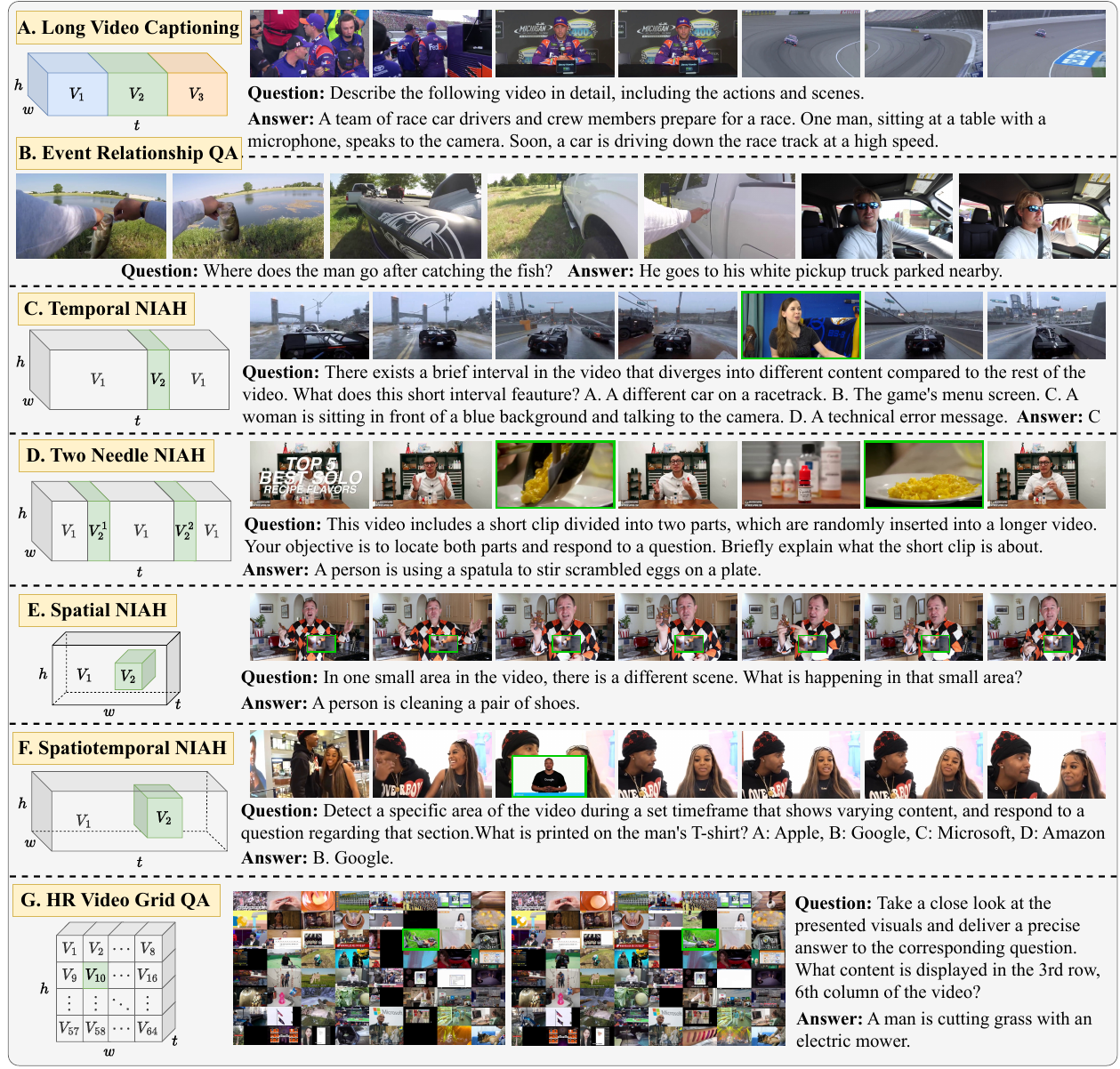}
  \vspace{-2em}
  \caption{Our proposed video augmentation and instruction-following data synthesis schemes for \dataset. Given input videos, We perform spatiotemporal video combinations to produce augmented video samples with longer duration and higher resolution.}
  \label{fig:main}
  \vspace{-5pt}
\end{figure*}
\section{\dataset Dataset}
Our goal is to create high-quality video instruction-following data from existing video-caption datasets. Specifically, we aim to (1) generate extended or higher-resolution videos by spatially or temporally combining existing videos, and (2) produce high-quality question-answer pairs by leveraging existing video captions. Formally, given a set of candidate videos $\mathbf{V} = \{V_1, V_2, ..., V_N\}$ with captions $\mathbf{C} = \{C_1, C_2, ..., C_N\}$, we seek to create an augmented video $V^*$ and synthesize a QA pair $(q, a)$:
\begin{align}
\label{eq:data_synthesis}
    V^* = \Phi(\mathbf{V}), \quad \quad (q, a) = \Theta(\mathbf{C}).
\end{align}
For the video augmentation operator $\Phi$, we draw inspiration from data augmentation techniques used in image and video classification, such as CutMix \cite{yun2019cutmix} and VideoMix \cite{yun2020videomix}, to perform video mixing and combination. We use a highly capable language model Gemini-1.5-Pro \cite{team2023gemini} as $\Theta$ to generate synthetic QA pairs. As shown in Figure~\ref{fig:main}, our dataset contains a total of seven subsets, each featuring different video augmentation methods. In this section, we detail the dataset construction process for each subset.

\begin{table*}[h!]
\centering
\small
\caption{Statistics of our synthetic video instruction-following dataset. ``(N)'' and ``(H)'' corresponds to the ``needle'' (short or low-res videos) and the ``haystack'' (long or high-res videos) in NIAH subsets.}
\setlength\tabcolsep{4 pt}
\begin{tabular}{@{}lccccc@{}}
\toprule
Subset                        & Instruction Type    & Video Source                   & \#Videos & Avg. Duration & Avg. Resolution \\ \midrule
Long Video Captioning         & Video Captioning & Panda-70M \cite{chen2024panda}                     & 58,617   & 33.2s         & 1277$\times$720        \\
Event Relationship QA         & Freeform QA/MCQ & Panda-70M \cite{chen2024panda}                     & 56,854   & 33.4s         & 1278$\times$720        \\
Temporal NIAH                 & Freeform QA/MCQ & Panda-70M \cite{chen2024panda} (N), MiraData \cite{ju2024miradata} (H)    & 59,751   & 67.6s         & 640$\times$358         \\
Two Needle NIAH               & Freeform QA     & Panda-70M \cite{chen2024panda} (N), FineVideo \cite{Farré2024FineVideo} (H)   & 52,349   & 112.4s        & 591$\times$382         \\
Spatial NIAH                  & Freeform QA/MCQ & InternVid \cite{wang2023internvid} (N), OpenVid-1M \cite{nan2024openvid} (H) & 59,978   & 9.9s          & 1726$\times$971        \\
Spatiotemporal NIAH           & Freeform QA/MCQ & OpenVid-1M \cite{nan2024openvid} (N), FineVideo \cite{Farré2024FineVideo} (H)     & 56,494   & 89.9s         & 591$\times$383         \\
HR Video Grid QA              & Freeform QA/MCQ & InternVid \cite{wang2023internvid}                & 59,901   & 3s            & 1920$\times$1080       \\ \midrule
\dataset                       & -                & -                              & 403,944  & 48.6s         & 1160$\times$666        \\ \bottomrule
\end{tabular}
\label{tab:data_stats}
\end{table*}

\subsection{Long Video Captioning \& Event QA} Our first approach to synthesizing long video instruction data is to generate longer videos by temporally concatenating multiple short clips, as illustrated by Figure~\ref{fig:main}-A. We observe that public video-text datasets like InternVid \cite{wang2023internvid} and Panda-70M \cite{chen2024panda} often contain short clips that are drawn from longer videos, which can be combined to form extended sequences. To harvest long video data, we combine multiple short clips from the same source video, ensuring that the interval between them does not exceed five seconds. This preserves natural scenes and content transitions in the extended videos. Given sampled videos containing multiple short clips and their accompanying captions, we generate two types of instruction-following data using Gemini:
\begin{itemize}
    \item [1.] \textbf{Long Video Captioning}: given the captions for each short clip, we prompt Gemini to generate a longer caption describing the entire video. This task is designed to enhance the summarization ability of video LMMs for input videos with longer durations.
    \item [2.] \textbf{Event Relationship QA}: we let Gemini generate freeform or multiple choice questions related to the order of the events based on the short clip captions. Figure~\ref{fig:main}-B shows an example of such instruction data. These QA pairs focus on improving video LMM's ability to recognize action and event sequences.
\end{itemize}

\subsection{Video Needle-in-a-haystack (NIAH) QA} 
To effectively understand and reason over long or high-resolution videos, video LMMs must learn to accurately retrieve relevant information from this long sequence of video tokens, i.e. finding ``needles in a haystack''. NIAH experiments have been widely adopted to evaluate LLMs \cite{Kamradt_NeedleInAHaystack} and have also been adapted to LMM evaluations \cite{wang2024needle, wang2024multimodal}. In this approach, we propose to spatially or temporally combine videos to form various NIAH QA data:
\begin{itemize}
    \item [1.] \textbf{Temporal NIAH}: as shown in Figure~\ref{fig:main}-C, this approach randomly inserts a short clip within a longer video, creating a ``temporal needle'' in the sequence. Based on this combined video, our instruction data contain questions that ask video LMMs to describe the content of the needle video. This task challenges video LMMs to locate the relevant needle tokens within a long video and accurately summarize their content.
    \item [2.] \textbf{Two Needle NIAH}: we consider a variant of Temporal NIAH, where a short video clip is split into two parts and inserted at different timestamps within a longer clip. Video LMMs must summarize the short clip's content by locating both ``needles'', challenging them to retrieve relevant information from multiple temporal locations within the video sequence.
    \item [3.] \textbf{Spatial NIAH}: recognizing local and small objects in high-resolution videos is essential for video LMMs, yet obtaining suitable training data for high-resolution videos with detailed captions or QA pairs is challenging. We propose to generate spatial NIAH data to simulate such training data. As depicted in Figure~\ref{fig:main}-E, we overlay a small and low-resolution video onto a high-resolution video at a random position, and then ask a question related to this small video. This approach focuses our QA data on a specific "needle" region, forcing video LMMs to extract local information from high-resolution videos.
    \item [4.] \textbf{Spatiotemporal NIAH}: finally, we generate synthetic training data by integrating spatial and temporal NIAH. As shown in Figure~\ref{fig:main}-F, we embed a low-resolution, short-duration ``needle'' video within a longer, high-resolution video at a random spatial position and timestamp. This combined setup encourages video LMMs to understand video contents across both spatial and temporal dimensions, fostering a more comprehensive understanding of complex video inputs.
\end{itemize}
For all four NIAH variants, we obtain freeform QA pairs by prompting Gemini using the captions of the ``needle'' videos. We further convert half of these QA pairs into multiple-choice questions by prompting Gemini to generate incorrect distractors derived from the captions of the ``haystack'' videos. This method ensures the distractor options are contextually related to the haystack content. Consequently, if the model fails to identify the needle video correctly, it is more likely to choose these distractor options, increasing the challenge and rigour of the NIAH tasks.

\subsection{High-Resolution Video Grid QA} 
In this task, we explore the idea of generating high-resolution video instruction data by combining multiple low-resolution videos. From a large collection of low-resolution video clips, we randomly sample 64 videos and arrange them in a $8\times 8$ grid (\textit{c.f.} Figure~\ref{fig:main}-G). Each video is resized to $240\times 135$, resulting in a combined video resolution of $1920\times 1080$. We then randomly select a cell at row $i$ and column $j$ in the video grid and synthesize a question about the content in this specific cell. This design enhances video LMM's ability to understand high-resolution videos by requiring it to locate the correct cell based on indices and accurately interpret the content within that small area. For MCQ data, we randomly select other cells within the grid and use their captions to create distractor options.

\subsection{Summary of Video Instruction-Following Data}
Table~\ref{tab:data_stats} provides a summary of our video instruction-following dataset. Our dataset comprises $\sim$400K entries, with each long video over 30 seconds and each high-resolution video at least 960p. A key advantage of our data synthesis pipeline is that our QA synthesis process (\textit{c.f.} Equation~\ref{eq:data_synthesis}) requires only text processing via the Gemini API, without needing Gemini's multimodal functionality. This makes our approach significantly more cost-efficient compared to existing methods \cite{chen2024sharegpt4video, zhang2024video}. We leverage five datasets: Panda-70M \cite{chen2024panda}, MiraData \cite{ju2024miradata}, FineVideo \cite{Farré2024FineVideo}, InternVid \cite{wang2023internvid} and OpenVid-1M \cite{nan2024openvid} to generate synthetic video instruction data. However, our method also applies to any raw video source by first using an off-the-shelf video captioning model to generate simple captions, followed by our pipeline to produce high-quality video instruction data.
\begin{table*}[]
\centering
\small
\caption{Comparisons between baseline models and \model-finetuned models on long/short video understanding benchmarks. The best results among open-source models are bolded. $\Delta$ denotes the performance differences before and after finetuning on \dataset.}
\setlength\tabcolsep{4.5 pt}
\begin{tabular}{@{}lcccccccccc@{}}
\toprule
\multicolumn{1}{l|}{}                                        & \multicolumn{1}{l|}{}                       & \multicolumn{7}{c|}{Long Video Understanding}                                                                                                                                                                                                                                                                                                                      & \multicolumn{2}{c}{Short Video Understanding}                               \\ \cmidrule(l){3-11} 
\multicolumn{1}{l|}{}                                        & \multicolumn{1}{l|}{}                       & \multicolumn{4}{c|}{Video-MME w/o subtitles}                                                                                                                                   & \multicolumn{1}{c|}{MLVU}                                 & \multicolumn{1}{c|}{LVBench}                              & \multicolumn{1}{c|}{LongVideoBench}                       & MVBench                              & NExT-QA                              \\ \cmidrule(l){3-11} 
\multicolumn{1}{l|}{\multirow{-3}{*}{Models}}                & \multicolumn{1}{l|}{\multirow{-3}{*}{Size}} & avg                                  & short                                & medium                               & \multicolumn{1}{c|}{long}                                 & \multicolumn{1}{c|}{m-avg}                                & \multicolumn{1}{c|}{test}                                 & \multicolumn{1}{c|}{val}                                  & test                                 & mc                                   \\ \midrule
\rowcolor{gray}
\multicolumn{11}{c}{\textit{Proprietary Models}}                                                                                                                                                                                                                                                                                                                                                                                                                                                                                                              \\
\multicolumn{1}{l|}{GPT-4V \cite{achiam2023gpt}}                                  & \multicolumn{1}{c|}{-}                      & 59.9                                 & 70.5                                 & 55.8                                 & \multicolumn{1}{c|}{53.5}                                 & \multicolumn{1}{c|}{49.2}                                 & \multicolumn{1}{c|}{-}                                    & \multicolumn{1}{c|}{59.1}                                 & 43.5                                 & -                                    \\
\multicolumn{1}{l|}{GPT-4o \cite{OpenAI_GPT4o}}                                  & \multicolumn{1}{c|}{-}                      & 71.9                                 & 80.0                                 & 70.3                                 & \multicolumn{1}{c|}{65.3}                                 & \multicolumn{1}{c|}{64.6}                                 & \multicolumn{1}{c|}{34.7}                                 & \multicolumn{1}{c|}{66.7}                                 & -                                    & 76.0                                 \\
\multicolumn{1}{l|}{Gemini-1.5-Pro \cite{team2023gemini}}                          & \multicolumn{1}{c|}{-}                      & 75.0                                 & 81.7                                 & 74.3                                 & \multicolumn{1}{c|}{67.4}                                 & \multicolumn{1}{c|}{-}                                    & \multicolumn{1}{c|}{33.1}                                 & \multicolumn{1}{c|}{64.0}                                 & -                                    & -                                    \\ \midrule
\rowcolor{gray}
\multicolumn{11}{c}{\textit{Open-source Models}}                                                                                                                                                                                                                                                                                                                                                                                                                                                                                                              \\
\multicolumn{1}{l|}{VideoChat2 \cite{li2024mvbench}}                              & \multicolumn{1}{c|}{7B}                     & 39.5                                 & 48.3                                 & 37.0                                 & \multicolumn{1}{c|}{33.2}                                 & \multicolumn{1}{c|}{47.9}                                 & \multicolumn{1}{c|}{-}                                    & \multicolumn{1}{c|}{39.3}                                 & 51.9                                 & \textbf{78.6}                        \\
\multicolumn{1}{l|}{LLaMA-VID \cite{li2025llama}}                               & \multicolumn{1}{c|}{7B}                     & -                                    & -                                    & -                                    & \multicolumn{1}{c|}{-}                                    & \multicolumn{1}{c|}{33.2}                                 & \multicolumn{1}{c|}{23.9}                                 & \multicolumn{1}{c|}{-}                                    & 41.3                                 & -                                    \\
\multicolumn{1}{l|}{ST-LLM \cite{liu2024large}}                              & \multicolumn{1}{c|}{7B}                    & 37.9                                 & 45.7                                 & 36.8                                 & \multicolumn{1}{c|}{31.3}                                 & \multicolumn{1}{c|}{-}                                 & \multicolumn{1}{c|}{-}                                    & \multicolumn{1}{c|}{-}                                    & 54.9                        & -                                    \\
\multicolumn{1}{l|}{ShareGPT4Video \cite{chen2024sharegpt4video}}                          & \multicolumn{1}{c|}{7B}                     & 39.9                                 & 48.3                                 & 36.3                                 & \multicolumn{1}{c|}{35.0}                                 & \multicolumn{1}{c|}{46.4}                                 & \multicolumn{1}{c|}{-}                                    & \multicolumn{1}{c|}{39.7}                                 & 51.2                                 & -                                    \\
\multicolumn{1}{l|}{LongVILA \cite{xue2024longvila}}                                & \multicolumn{1}{c|}{7B}                     & 50.5                                 & 61.8                                 & 50.4                                 & \multicolumn{1}{c|}{46.2}                                 & \multicolumn{1}{c|}{-}                                    & \multicolumn{1}{c|}{-}                                    & \multicolumn{1}{c|}{-}                                    & -                                    & -                                    \\
\multicolumn{1}{l|}{LongLLaVA \cite{wang2024longllava}}                               & \multicolumn{1}{c|}{7B}                     & 52.9                                 & 61.9                                 & 51.4                                 & \multicolumn{1}{c|}{45.4}                                 & \multicolumn{1}{c|}{-}                                    & \multicolumn{1}{c|}{-}                                    & \multicolumn{1}{c|}{-}                                    & 54.6                                 & -                                    \\
\multicolumn{1}{l|}{Video-XL \cite{shu2024video}}                                & \multicolumn{1}{c|}{7B}                     & \textbf{55.5}                        & 64.0                                 & \textbf{53.2}                        & \multicolumn{1}{c|}{\textbf{49.2}}                        & \multicolumn{1}{c|}{\textbf{64.9}}                        & \multicolumn{1}{c|}{-}                                    & \multicolumn{1}{c|}{49.5}                                 & \textbf{55.3}                                 & 77.2                                 \\ \midrule
\multicolumn{1}{l|}{VideoLLaVA \cite{lin2023video}}                              & \multicolumn{1}{c|}{7B}                     & 39.9                                 & 45.3                                 & 38.0                                 & \multicolumn{1}{c|}{36.2}                                 & \multicolumn{1}{c|}{47.3}                                 & \multicolumn{1}{c|}{29.3}                                 & \multicolumn{1}{c|}{39.9}                                 & 43.8                                 & 61.8                                 \\
\multicolumn{1}{l|}{\model-VideoLLaVA}        & \multicolumn{1}{c|}{7B}                     & 43.7                                 & 48.2                                 & 43.9                                 & \multicolumn{1}{c|}{38.9}                                 & \multicolumn{1}{c|}{49.5}                                 & \multicolumn{1}{c|}{33.8}                                 & \multicolumn{1}{c|}{42.3}                                 & 47.2                                 & 63.0                                 \\
\rowcolor{LightCyan}
\multicolumn{1}{l|}{$\Delta$ - VideoLLaVA}      & \multicolumn{1}{c|}{}                       & {\color[HTML]{009901} \textbf{+3.8}} & {\color[HTML]{009901} \textbf{+2.9}} & {\color[HTML]{009901} \textbf{+5.9}} & \multicolumn{1}{c|}{{\color[HTML]{009901} \textbf{+2.7}}} & \multicolumn{1}{c|}{{\color[HTML]{009901} \textbf{+2.2}}} & \multicolumn{1}{c|}{{\color[HTML]{009901} \textbf{+4.5}}} & \multicolumn{1}{c|}{{\color[HTML]{009901} \textbf{+2.4}}} & {\color[HTML]{009901} \textbf{+3.4}} & {\color[HTML]{009901} \textbf{+1.2}} \\ \midrule
\multicolumn{1}{l|}{Mantis-Idefics2 \cite{jiang2024mantis}}                         & \multicolumn{1}{c|}{8B}                     & 45.4                                 & 55.9                                 & 43.0                                 & \multicolumn{1}{c|}{37.2}                                 & \multicolumn{1}{c|}{49.4}                                 & \multicolumn{1}{c|}{35.0}                                 & \multicolumn{1}{c|}{45.8}                                & 51.4                                 & 75.8                                 \\
\multicolumn{1}{l|}{\model-Mantis}            & \multicolumn{1}{c|}{8B}                     & 48.2                                 & 58.4                                 & 46.7                                 & \multicolumn{1}{c|}{39.6}                                 & \multicolumn{1}{c|}{55.5}                                 & \multicolumn{1}{c|}{36.4}                                 & \multicolumn{1}{c|}{49.1}                                 & 52.5                                 & 75.2                                 \\
\rowcolor{LightCyan}
\multicolumn{1}{l|}{$\Delta$ - Mantis-Idefics2} & \multicolumn{1}{c|}{}                       & {\color[HTML]{009901} \textbf{+2.8}} & {\color[HTML]{009901} \textbf{+2.5}} & {\color[HTML]{009901} \textbf{+3.7}} & \multicolumn{1}{c|}{{\color[HTML]{009901} \textbf{+2.4}}} & \multicolumn{1}{c|}{{\color[HTML]{009901} \textbf{+6.1}}} & \multicolumn{1}{c|}{{\color[HTML]{009901} \textbf{+1.4}}} & \multicolumn{1}{c|}{{\color[HTML]{009901} \textbf{+3.3}}} & {\color[HTML]{009901} \textbf{+1.1}} & {\color[HTML]{FE0000} \textbf{-0.6}} \\ \midrule
\multicolumn{1}{l|}{LongVA \cite{zhang2024long}}                                  & \multicolumn{1}{c|}{7B}                     & 52.4                                 & 61.4                                 & 50.9                                 & \multicolumn{1}{c|}{45.0}                                 & \multicolumn{1}{c|}{56.3}                                 & \multicolumn{1}{c|}{35.9}                                 & \multicolumn{1}{c|}{51.8}                                 & 49.2                                 & 68.3                                 \\
\multicolumn{1}{l|}{\model-LongVA}            & \multicolumn{1}{c|}{7B}                     & \textbf{55.5}                        & \textbf{66.0}                        & 53.1                                 & \multicolumn{1}{c|}{47.4}                                 & \multicolumn{1}{c|}{62.1}                                 & \multicolumn{1}{c|}{\textbf{39.0}}                        & \multicolumn{1}{c|}{\textbf{53.1}}                        & 51.1                                 & 69.3                                 \\
\rowcolor{LightCyan}
\multicolumn{1}{l|}{$\Delta$ - LongVA}          & \multicolumn{1}{c|}{}                       & {\color[HTML]{009901} \textbf{+3.1}} & {\color[HTML]{009901} \textbf{+4.6}} & {\color[HTML]{009901} \textbf{+2.2}} & \multicolumn{1}{c|}{{\color[HTML]{009901} \textbf{+2.4}}} & \multicolumn{1}{c|}{{\color[HTML]{009901} \textbf{+5.8}}} & \multicolumn{1}{c|}{{\color[HTML]{009901} \textbf{+3.1}}} & \multicolumn{1}{c|}{{\color[HTML]{009901} \textbf{+1.3}}} & {\color[HTML]{009901} \textbf{+1.9}} & {\color[HTML]{009901} \textbf{+1.0}} \\ \bottomrule
\end{tabular}
\label{tab:main_results}
\end{table*}

\vspace{-1em}
\section{Evaluation: \hrbench}
\label{sec:hrbench}
We observe that existing video understanding benchmarks are inadequate for accurately assessing the ability of video LMMs to understand high-resolution videos, especially the details inside the videos. Prior benchmarks mainly consist of low-resolution videos. More recent benchmarks focus on evaluating the long video understanding capability of video LMMs, which contain questions that typically pertain to a short segment in the long video. As a result, a model's high-resolution video understanding performance can be undermined if it struggles to sample or retrieve the relevant frames from a lengthy video sequence.

To address this gap, we introduce \hrbench, a comprehensive benchmark with 200 multiple-choice questions designed to assess video LMMs for high-resolution video understanding. \hrbench focuses on the perception and understanding of small regions and subtle actions in the video. Our test videos are at least 1080p and contain 10 different video types collected with real-world applications in mind. For example, key applications of high-resolution video understanding include autonomous driving and video surveillance. We correspondingly collect POV driving videos and CCTV footage for the benchmark. Our benchmark consists of 10 types of questions, all of which are manually annotated and can be broadly categorized into object and action-related tasks. Further details can be found in the Appendix. The question types are:
\begin{itemize}
    \item Object-related Tasks: Object Counting, OCR problem, Object Recognition, Entity Recognition, Object Property Recognition, Object Status Change Recognition.
    \item Action-related Tasks: Action Recognition, Moving Direction Identification, Interaction Detection, Temporal Sequence Recognition.
\end{itemize}

\begin{table*}[]
\caption{Quantitative results on \hrbench and open-ended video QA benchmarks. ``acc.'' represents accuracy.}
\centering
\small
\setlength\tabcolsep{7 pt}
\begin{tabular}{@{}l|ccc|cccccccc@{}}
\toprule
                                   & \multicolumn{3}{l|}{High-Res Video Understanding}                                                                     & \multicolumn{8}{c}{Open-Ended Video QA}                                                                                                                                                                                                                                                                                  \\ \cmidrule(l){2-12} 
                                   & \multicolumn{3}{c|}{\hrbench}                                                                                     & \multicolumn{2}{c}{MSVD-QA}                                                  & \multicolumn{2}{c}{MSRVTT-QA}                                                & \multicolumn{2}{c}{TGIF-QA}                                                  & \multicolumn{2}{c}{ActivityNet-QA}                                          \\ \cmidrule(l){2-12} 
\multirow{-3}{*}{Models}           & avg                                  & object                                & action                                & acc.                                  & score                                & acc.                                  & score                                & acc.                                  & score                                & acc.                                 & score                                \\ \midrule
VideoLLaVA \cite{lin2023video}                         & 32.5                                  & 36.0                                  & 27.9                                  & 60.3                                  & 3.7                                  & 42.1                                  & 3.0                                  & 63.5                                  & 3.8                                  & 48.6                                 & 3.3                                  \\
\model-VideoLLaVA   & 47.5                                  & 50.0                                  & 44.2                                  & 71.5                                  & 4.0                                  & 58.5                                  & 3.5                                  & 78.0                                  & 4.3                                  & 49.1                                 & 3.4                                  \\
\rowcolor{LightCyan}
$\Delta$ - VideoLLaVA & {\color[HTML]{009901} \textbf{+15.0}} & {\color[HTML]{009901} \textbf{+14.0}} & {\color[HTML]{009901} \textbf{+16.3}} & {\color[HTML]{009901} \textbf{+11.2}} & {\color[HTML]{009901} \textbf{+0.3}} & {\color[HTML]{009901} \textbf{+16.4}} & {\color[HTML]{009901} \textbf{+0.5}} & {\color[HTML]{009901} \textbf{+14.5}} & {\color[HTML]{009901} \textbf{+0.5}} & {\color[HTML]{009901} \textbf{+0.5}} & {\color[HTML]{009901} \textbf{+0.1}} \\ \midrule
Mantis-Idefics2 \cite{jiang2024mantis}                   & 48.5                                  & 50.9                                  & 45.4                                  & 57.4                                  & 3.5                                  & 34.9                                  & 2.7                                  & 65.7                                  & 3.8                                  & 46.5                                 & 3.1                                  \\
\model-Mantis       & 51.0                                  & 53.5                                  & 47.7                                  & 65.2                                  & 3.8                                  & 46.4                                  & 3.1                                  & 71.4                                  & 4.0                                  & 48.8                                 & 3.3                                  \\
\rowcolor{LightCyan}
$\Delta$ - Mantis     & {\color[HTML]{009901} \textbf{+2.5}}  & {\color[HTML]{009901} \textbf{+2.6}}  & {\color[HTML]{009901} \textbf{+2.3}}  & {\color[HTML]{009901} \textbf{+7.8}}  & {\color[HTML]{009901} \textbf{+0.3}} & {\color[HTML]{009901} \textbf{+11.5}} & {\color[HTML]{009901} \textbf{+0.4}} & {\color[HTML]{009901} \textbf{+5.7}}  & {\color[HTML]{009901} \textbf{+0.2}} & {\color[HTML]{009901} \textbf{+2.3}} & {\color[HTML]{009901} \textbf{+0.2}} \\ \midrule
LongVA \cite{zhang2024long}                         & 48.0                                  & 52.6                                  & 41.9                                  & 56.3                                  & 3.5                                  & 37.7                                  & 2.8                                  & 55.4                                  & 3.4                                  & 48.0                                 & 3.2                                  \\
\model-LongVA     & 50.0                                  & 56.1                                  & 41.9                                  & 61.0                                  & 3.7                                  & 42.5                                  & 3.0                                  & 67.5                                  & 3.9                                  & 51.8                                 & 3.4                                  \\
\rowcolor{LightCyan}
$\Delta$ - LongVA     & {\color[HTML]{009901} \textbf{+2.0}}  & {\color[HTML]{009901} \textbf{+3.5}}  & \textbf{+0.0}  & {\color[HTML]{009901} \textbf{+4.7}}  & {\color[HTML]{009901} \textbf{+0.2}} & {\color[HTML]{009901} \textbf{+4.8}}  & {\color[HTML]{009901} \textbf{+0.2}} & {\color[HTML]{009901} \textbf{+12.1}} & {\color[HTML]{009901} \textbf{+0.5}} & {\color[HTML]{009901} \textbf{+3.8}} & {\color[HTML]{009901} \textbf{+0.2}} \\ \bottomrule
\end{tabular}
\label{tab:hr_oe_qa}
\end{table*}
\section{Experimental Results}

\subsection{Evaluation Setup}
\label{sec:eval_setup}
To validate the effectiveness of \dataset, we finetune a diverse set of LMMs on our dataset. Specifically, we choose VideoLLaVA \cite{lin2023video}, Mantis-Idefics2 \cite{jiang2024mantis} and LongVA \cite{zhang2024long} as the base models because these models disclose details about their training dataset. Further details of these models can be found in the Appendix. We directly finetune LongVA and Mantis-Idefics2 on our dataset as they are mainly pretrained on image data. For VideoLLaVA, we additionally include 300K short video from VideoChat2-IT \cite{li2024mvbench} to preserve the model's short video understanding capability. We use 8 frames for VideoLLaVA, 24 frames for Mantis-Idefics2, and 64 frames for LongVA for finetuning and evaluation. The input resolution is fixed at 224$\times$224 for VideoLLaVA and 336$\times$336 for LongVA, while Mantis-Idefics2 supports dynamic resolutions up to 980$\times$980. The finetuned model is denoted as \model-[base model].

Our evaluation assesses video LMMs' capabilities in long and high-resolution video understanding. We test the finetuned models on four comprehensive long video understanding benchmarks: Video-MME \cite{fu2024video}, MLVU \cite{zhou2024mlvu}, LVBench \cite{wang2024lvbench} and LongVideoBench \cite{wu2024longvideobench}. For high-resolution video understanding, we utilize our \hrbench. Additionally, we report results on prior short video understanding benchmarks including MVBench \cite{li2024mvbench} and NExT-QA \cite{xiao2021next}, as well as open-ended video QA benchmarks such as MSVD-QA \cite{xu2017video}, MSRVTT-QA \cite{xu2017video}, TGIF-QA \cite{jang2017tgif} and ActivityNet-QA \cite{yu2019activitynet}.

\subsection{Quantitative Results}
\paragraph{Long Video Understanding} Quantitative results for long video understanding are shown in Table~\ref{tab:main_results}. All three models demonstrate a consistent performance boost across the evaluation benchmarks, with average improvements of 3.3\% for Video-MME, 4.7\% for MLVU, 3.0\% for LVBench, and 2.3\% for LongVideoBench. For Video-MME, our method yields the largest average improvement for medium-length questions, which correspond to videos ranging from 4 to 15 minutes. Given that most of the augmented videos in our dataset are under 3 minutes, this result suggests that our approach generalizes effectively to longer videos. One possible explanation for this observation is that longer videos often contain more scene cuts and transitions, which is aligned with our synthetic data where multiple clips with different scenes are concatenated. Our \model-LongVA achieves state-of-the-art performance on Video-MME, LVBench and LongVideoBench among open-source models, demonstrating the effectiveness of our approach.

\vspace{-1em}
\paragraph{High-Resolution Video Understanding}
We present the \hrbench evaluation results for \model-finetuned models and the baseline models in Table~\ref{tab:hr_oe_qa}. Among the baseline models, we observe that Mantis-Idefics2, with its higher input video resolution, outperforms LongVA on both object and action-related tasks, despite being considered weaker for long video understanding. This suggests that our \hrbench requires detailed comprehension of high-resolution videos, and simply adding more input frames during testing does not necessarily lead to enhanced model performance. Overall, we find that finetuning on our dataset enhances the high-resolution video understanding capabilities for all three models. Additionally, we observe that recognizing subtle actions remains more challenging than identifying objects and their properties, suggesting that there remains significant potential for improvement in high-resolution video understanding for video LMMs.

\vspace{-1em}
\paragraph{Short Video Understanding}
We further validate our proposed method on several prior short video understanding benchmarks. According to Table~\ref{tab:main_results}, all three \model-finetuned models show improvement on both MVBench and NExT-QA benchmarks except Mantis. This result is likely due to NExT-QA’s training split being included as part of the in-domain training data for Mantis-Idefics2. Consequently, finetuning on our out-of-domain data causes a slight decrease in performance. While our results are slightly worse than Video-XL \cite{shu2024video}, we note that their pretraining and instruction-tuning datasets are larger ($>$2.7M training samples). Here, we focus more on the relative improvement of our method over the baseline models.

Since the benchmarks mentioned above consist solely of multiple-choice questions, we also include several open-ended QA benchmarks to verify the text generation capabilities of the finetuned models. Following Video-ChatGPT \cite{maaz2023video}, we use GPT-3.5-Turbo \cite{OpenAI_ChatGPT_Website} to evaluate the accuracy of the responses and rate their quality on a scale of 1 to 5. As shown in Table~\ref{tab:hr_oe_qa}, finetuning on our dataset improves both accuracy and generation quality across all four benchmarks for the \model-finetuned models. This demonstrates that our dataset can enhance the generation quality of LMMs.

\subsection{Ablation Study}
To verify that each subset in \dataset contributes to the performance gain of long or high-resolution video understanding, we conduct an ablation study and train several Mantis-Idefics2 models by excluding each subset from the training data. In addition, we add the same amount of video training data from VideoChat2-IT \cite{li2024mvbench} to each of the ablation experiments to ensure the total number of training examples does not change. The results on Video-MME and \hrbench are shown in Table~\ref{tab:ablation}. We observe that excluding the long video subsets decreases the model performances on Video-MME. Similarly, excluding high-resolution video subsets leads to a drop in scores on \hrbench. These observations indicate that all seven video augmentation methods are effective at their corresponding tasks. We also report results for a Mantis-Idefics2 model without our proposed video augmentations by using the original videos directly for training before combining them with others (e.g., using the needle video in NIAH tasks to train the model directly). According to the last row in Table~\ref{tab:ablation}, this yields very low scores on Video-MME and \hrbench, highlighting the necessity of our augmentations to enhance video instruction data quality.

Furthermore, previous studies \cite{li2024llava, zhang2024long} have shown that training on high-resolution image data yields strong long video understanding models. Our ablation study supports this conclusion and demonstrates that the proposed high-resolution video augmentation methods enhance the performances of long video understanding tasks. In addition, our results indicate that the reverse is also true: our long video augmentation methods also improve the performance of high-resolution video understanding. We believe this transferability between long and high-resolution video understanding capabilities is because both types of inputs convert to a long sequence of video tokens in video LMM’s embedding space. As a result, identifying short moments from a long video and recognizing small regions in a high-resolution video both correspond to the task of finding key information needles in a haystack of video tokens.

\begin{table}[]
\centering
\small
\caption{Ablation study results for \model-Mantis. Each ``w/o [Subset]'' denotes a Mantis-Idefics2 model finetuned on a modified \dataset by replacing the corresponding subset with the same amount of training examples from VideoChat2-IT \cite{li2024mvbench}.}
\setlength\tabcolsep{2.5 pt}
\begin{tabular}{@{}lcc@{}}
\toprule
\multirow{2}{*}{Models}   & Video-MME & \hrbench \\ \cmidrule(l){2-3} 
                          & w/o sub. avg                    & avg         \\ \midrule
\model-Mantis             & \textbf{48.2}           & \textbf{51.0}\\
\midrule
w/o Long Video Captioning & 47.9                    & 48.0         \\
w/o Event Relationship QA & 47.7                    & 49.5         \\
w/o Temporal NIAH         & 47.5                    & 48.0         \\
w/o Two Needle NIAH       & 48.1                    & 50.5         \\
w/o Spatial NIAH          & 47.2                    & 47.5         \\
w/o Spatiotemporal NIAH   & 47.7                    & 50.0         \\
w/o HR Video Grid QA      & 47.8                    & 48.0           \\  \midrule
w/o Video Augmentation   & 45.7                    & 44.5         \\ \bottomrule
\end{tabular}
\label{tab:ablation}
\vspace{-1em}
\end{table}

\begin{figure*}[t!]
  \centering
  \includegraphics[width=1.0\textwidth]{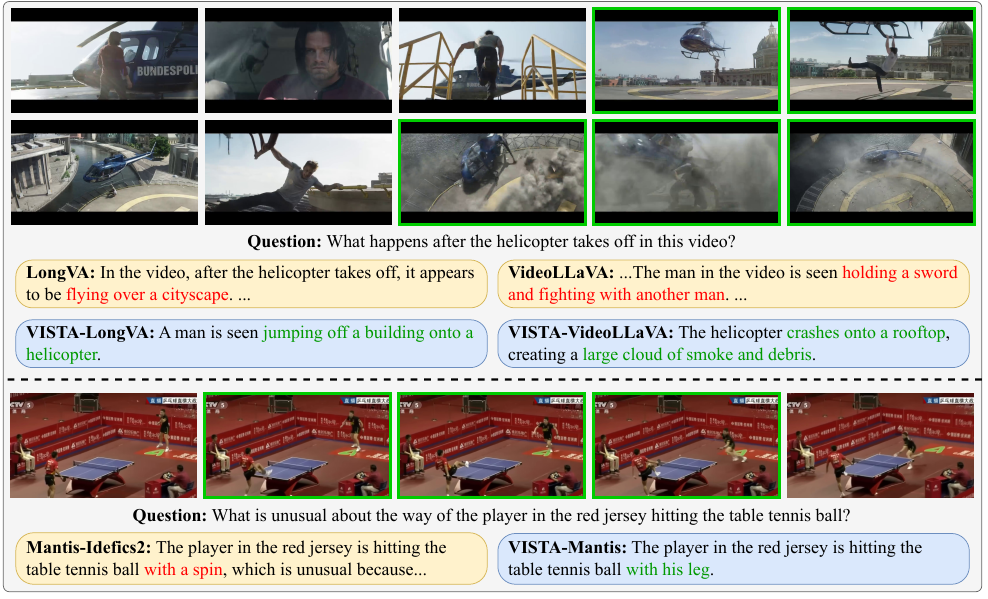}
  \vspace{-2em}
  \caption{Qualitative comparisons between the baseline models and our \model-finetuned models. \textcolor{red}{Red text} indicates hallucinations or incorrect responses, while \textcolor{deepgreen}{green text} highlights the correct responses that correspond accurately to the video content.}
  \label{fig:case_study}
  \vspace{-5pt}
\end{figure*}

\subsection{Case Study}
To qualitatively evaluate our method, we compare the generated responses of \model-finetuned models with the baseline models, as illustrated in Figure~\ref{fig:case_study}. The ``helicopter'' example involves recognizing event orders within a long video sequence. The baseline models LongVA and VideoLLaVA exhibit hallucination issues, while our \model-LongVA describes the action accurately. Due to the sparse sampling rate of VideoLLaVA (8 frames per video), \model-VideoLLaVA misses frames related to the action immediately following ``the helicopter takes off''. Nevertheless, it successfully describes the later event, ``the helicopter crashes onto a rooftop'', which the vanilla VideoLLaVA fails to recognize. The ``table tennis'' example challenges video LMMs to identify a subtle and unusual action within a high-resolution video. As demonstrated in the second part of Figure~\ref{fig:case_study}, Mantis-Idefics2 fails to recognize this action, instead generating a response based on common knowledge about table tennis. Conversely, \model-Mantis correctly identifies the unusual action that the player hits the table tennis ball with his leg.

\section{Related Work}
\label{sec:related_work}
\paragraph{LMMs for Video Understanding}
Recent research has advanced the ability of large multimodal models (LMMs) to process video inputs and generate natural language responses. Earlier works on video LMMs such as VideoChatGPT \cite{maaz2023video} and Video-LLaVA \cite{lin2023video} employ video encoders that are restricted to a limited number of low-resolution frames (e.g. $224\times224$, eight frames). To enable the processing of longer and higher-resolution videos, several studies \cite{zhang2023video, li2023videochat, li2024mvbench, song2024moviechat, ren2024timechat, fei2024video, li2025llama, liu2024kangaroo} have developed token compression and pooling methods to manage the sequence length of video tokens. Video-LLaMA \cite{zhang2023video}, VideoChat \cite{li2023videochat} and VideoChat2 \cite{li2024mvbench} utilize Q-Former \cite{li2023blip} in the visual encoder to compress video tokens into a few learnable queries. Video-CCAM \cite{fei2024video} further applies causal cross-attention masks within the video Q-Former to better capture the temporal order of the video frames. In addition to using Q-Former, VideoLLaMA 2 \cite{cheng2024videollama} designs a 3D convolution-based STC Connector for efficient token compression. LongVA \cite{zhang2024long} alternatively increases the maximum number of tokens the model can receive by training a long-context LLM with a 128K context length, allowing the model to handle longer video sequences.

\vspace{-1em}
\paragraph{Video Language Datasets}
Previous efforts on building video-text datasets have typically focused on a specialized domain, such as video retrieval \cite{xu2016msr, anne2017localizing, krishna2017dense}, video question answering \cite{xiao2021next, yu2019activitynet, xu2017video} and action recognition \cite{soomro2012ucf101, carreira2017quo, goyal2017something}. VideoChat2 \cite{li2024mvbench} creates a large-scale video instruction tuning dataset by combining multiple specialized datasets and rewriting instructions using ChatGPT \cite{OpenAI_ChatGPT_Website}. However, the quality of the videos is suboptimal. More recently, LLaVA-Hound \cite{zhang2024direct} and ShareGPT4Video \cite{chen2024sharegpt4video} generate high-quality video captioning and open-ended QA data using GPT-4V \cite{achiam2023gpt}. Nonetheless, the video sources in these datasets often have limited motion degree and variety.

A concurrent study to ours is the LLaVA-Video-178K dataset \cite{zhang2024video}, which contains 178K high-quality video data and 1.3M instruction-following samples annotated by GPT-4o \cite{OpenAI_GPT4o}. However, this dataset relies on a costly annotation process that involves feeding in video frames to GPT-4o at one fps and generating three levels of video descriptions. In contrast, our work emphasizes the synthesis of high-quality video data from existing short or low-resolution videos, offering a more scalable and cost-effective solution for generating synthetic video language data.

\vspace{-1em}
\paragraph{Video Understanding Benchmarks}
The advancement of video LMMs has also driven the creation of various video understanding benchmarks. MVBench \cite{li2024mvbench} introduces a comprehensive video LMM benchmark by regenerating question-answer pairs based on various existing benchmarks. TempCompass \cite{liu2024tempcompass} and VideoVista \cite{li2024videovista} evaluate the temporal reasoning and understanding capabilities of video LMMs. Recently, new benchmarks have emerged to assess video LMMs' ability to comprehend extremely long videos, including Video-MME \cite{fu2024video}, MLVU \cite{zhou2024mlvu}, LVBench \cite{wang2024lvbench} and LongVideoBench \cite{wu2024longvideobench}. Despite these advancements, we note that no existing benchmark specifically evaluates video LMMs' ability to understand high-resolution videos, which is a crucial task in video understanding that has significant applications in fields such as autonomous driving and sports analysis. In this study, we aim to address this gap by collecting a comprehensive benchmark for high-resolution video understanding targeting small objects and localized actions within videos.
\section{Conclusion}
We presented \model, a video augmentation framework to enhance long-duration and high-resolution video understanding by generating high-quality synthetic video instruction-following data from existing video-caption datasets. \model performs spatiotemporal video combination over input videos to create new video samples and utilizes video captions for instruction synthesis. Through extensive experiments on various long video understanding benchmarks and our proposed \hrbench, we demonstrated that \dataset consistently improves the performance of three different video LMMs. For future work, we aim to design and explore additional video augmentation methods to further strengthen our approach's robustness.
{
    \small
    \bibliographystyle{ieeenat_fullname}
    \bibliography{ref}

\begin{thebibliography}{67}
\providecommand{\natexlab}[1]{#1}
\providecommand{\url}[1]{\texttt{#1}}
\expandafter\ifx\csname urlstyle\endcsname\relax
  \providecommand{\doi}[1]{doi: #1}\else
  \providecommand{\doi}{doi: \begingroup \urlstyle{rm}\Url}\fi

\bibitem[Achiam et~al.(2023)Achiam, Adler, Agarwal, Ahmad, Akkaya, Aleman, Almeida, Altenschmidt, Altman, Anadkat, et~al.]{achiam2023gpt}
Josh Achiam, Steven Adler, Sandhini Agarwal, Lama Ahmad, Ilge Akkaya, Florencia~Leoni Aleman, Diogo Almeida, Janko Altenschmidt, Sam Altman, Shyamal Anadkat, et~al.
\newblock Gpt-4 technical report.
\newblock \emph{arXiv preprint arXiv:2303.08774}, 2023.

\bibitem[Anne~Hendricks et~al.(2017)Anne~Hendricks, Wang, Shechtman, Sivic, Darrell, and Russell]{anne2017localizing}
Lisa Anne~Hendricks, Oliver Wang, Eli Shechtman, Josef Sivic, Trevor Darrell, and Bryan Russell.
\newblock Localizing moments in video with natural language.
\newblock In \emph{Proceedings of the IEEE international conference on computer vision}, pages 5803--5812, 2017.

\bibitem[Carreira and Zisserman(2017)]{carreira2017quo}
Joao Carreira and Andrew Zisserman.
\newblock Quo vadis, action recognition? a new model and the kinetics dataset.
\newblock In \emph{proceedings of the IEEE Conference on Computer Vision and Pattern Recognition}, pages 6299--6308, 2017.

\bibitem[Chen et~al.(2024{\natexlab{a}})Chen, Wei, Li, Dong, Zhang, Zang, Chen, Duan, Lin, Tang, et~al.]{chen2024sharegpt4video}
Lin Chen, Xilin Wei, Jinsong Li, Xiaoyi Dong, Pan Zhang, Yuhang Zang, Zehui Chen, Haodong Duan, Bin Lin, Zhenyu Tang, et~al.
\newblock Sharegpt4video: Improving video understanding and generation with better captions.
\newblock \emph{arXiv preprint arXiv:2406.04325}, 2024{\natexlab{a}}.

\bibitem[Chen et~al.(2024{\natexlab{b}})Chen, Siarohin, Menapace, Deyneka, Chao, Jeon, Fang, Lee, Ren, Yang, et~al.]{chen2024panda}
Tsai-Shien Chen, Aliaksandr Siarohin, Willi Menapace, Ekaterina Deyneka, Hsiang-wei Chao, Byung~Eun Jeon, Yuwei Fang, Hsin-Ying Lee, Jian Ren, Ming-Hsuan Yang, et~al.
\newblock Panda-70m: Captioning 70m videos with multiple cross-modality teachers.
\newblock In \emph{Proceedings of the IEEE/CVF Conference on Computer Vision and Pattern Recognition}, pages 13320--13331, 2024{\natexlab{b}}.

\bibitem[Cheng et~al.(2024)Cheng, Leng, Zhang, Xin, Li, Chen, Zhu, Zhang, Luo, Zhao, et~al.]{cheng2024videollama}
Zesen Cheng, Sicong Leng, Hang Zhang, Yifei Xin, Xin Li, Guanzheng Chen, Yongxin Zhu, Wenqi Zhang, Ziyang Luo, Deli Zhao, et~al.
\newblock Videollama 2: Advancing spatial-temporal modeling and audio understanding in video-llms.
\newblock \emph{arXiv preprint arXiv:2406.07476}, 2024.

\bibitem[Dao(2023)]{dao2023flashattention}
Tri Dao.
\newblock Flashattention-2: Faster attention with better parallelism and work partitioning.
\newblock \emph{arXiv preprint arXiv:2307.08691}, 2023.

\bibitem[Farré et~al.(2024)Farré, Marafioti, Tunstall, Von~Werra, and Wolf]{Farré2024FineVideo}
Miquel Farré, Andi Marafioti, Lewis Tunstall, Leandro Von~Werra, and Thomas Wolf.
\newblock Finevideo.
\newblock \url{https://huggingface.co/datasets/HuggingFaceFV/finevideo}, 2024.

\bibitem[Fei et~al.(2024)Fei, Li, Deng, Wang, Liu, and Wang]{fei2024video}
Jiajun Fei, Dian Li, Zhidong Deng, Zekun Wang, Gang Liu, and Hui Wang.
\newblock Video-ccam: Enhancing video-language understanding with causal cross-attention masks for short and long videos.
\newblock \emph{arXiv preprint arXiv:2408.14023}, 2024.

\bibitem[Fu et~al.(2024)Fu, Dai, Luo, Li, Ren, Zhang, Wang, Zhou, Shen, Zhang, et~al.]{fu2024video}
Chaoyou Fu, Yuhan Dai, Yondong Luo, Lei Li, Shuhuai Ren, Renrui Zhang, Zihan Wang, Chenyu Zhou, Yunhang Shen, Mengdan Zhang, et~al.
\newblock Video-mme: The first-ever comprehensive evaluation benchmark of multi-modal llms in video analysis.
\newblock \emph{arXiv preprint arXiv:2405.21075}, 2024.

\bibitem[Goyal et~al.(2017)Goyal, Ebrahimi~Kahou, Michalski, Materzynska, Westphal, Kim, Haenel, Fruend, Yianilos, Mueller-Freitag, et~al.]{goyal2017something}
Raghav Goyal, Samira Ebrahimi~Kahou, Vincent Michalski, Joanna Materzynska, Susanne Westphal, Heuna Kim, Valentin Haenel, Ingo Fruend, Peter Yianilos, Moritz Mueller-Freitag, et~al.
\newblock The" something something" video database for learning and evaluating visual common sense.
\newblock In \emph{Proceedings of the IEEE international conference on computer vision}, pages 5842--5850, 2017.

\bibitem[Jang et~al.(2017)Jang, Song, Yu, Kim, and Kim]{jang2017tgif}
Yunseok Jang, Yale Song, Youngjae Yu, Youngjin Kim, and Gunhee Kim.
\newblock Tgif-qa: Toward spatio-temporal reasoning in visual question answering.
\newblock In \emph{Proceedings of the IEEE conference on computer vision and pattern recognition}, pages 2758--2766, 2017.

\bibitem[Jiang et~al.(2024)Jiang, He, Zeng, Wei, Ku, Liu, and Chen]{jiang2024mantis}
Dongfu Jiang, Xuan He, Huaye Zeng, Cong Wei, Max Ku, Qian Liu, and Wenhu Chen.
\newblock Mantis: Interleaved multi-image instruction tuning.
\newblock \emph{arXiv preprint arXiv:2405.01483}, 2024.

\bibitem[Ju et~al.(2024)Ju, Gao, Zhang, Yuan, Wang, Zeng, Xiong, Xu, and Shan]{ju2024miradata}
Xuan Ju, Yiming Gao, Zhaoyang Zhang, Ziyang Yuan, Xintao Wang, Ailing Zeng, Yu Xiong, Qiang Xu, and Ying Shan.
\newblock Miradata: A large-scale video dataset with long durations and structured captions.
\newblock \emph{arXiv preprint arXiv:2407.06358}, 2024.

\bibitem[Kamradt(2024)]{Kamradt_NeedleInAHaystack}
Garrett Kamradt.
\newblock Llmtest\_needleinahaystack.
\newblock \url{https://github.com/gkamradt/LLMTest_NeedleInAHaystack}, 2024.
\newblock Accessed: 2024-10-24.

\bibitem[Kingma(2014)]{kingma2014adam}
Diederik~P Kingma.
\newblock Adam: A method for stochastic optimization.
\newblock \emph{arXiv preprint arXiv:1412.6980}, 2014.

\bibitem[Krishna et~al.(2017)Krishna, Hata, Ren, Fei-Fei, and Carlos~Niebles]{krishna2017dense}
Ranjay Krishna, Kenji Hata, Frederic Ren, Li Fei-Fei, and Juan Carlos~Niebles.
\newblock Dense-captioning events in videos.
\newblock In \emph{Proceedings of the IEEE international conference on computer vision}, pages 706--715, 2017.

\bibitem[Lauren{\c{c}}on et~al.(2024)Lauren{\c{c}}on, Tronchon, Cord, and Sanh]{laurenccon2024matters}
Hugo Lauren{\c{c}}on, L{\'e}o Tronchon, Matthieu Cord, and Victor Sanh.
\newblock What matters when building vision-language models?
\newblock \emph{arXiv preprint arXiv:2405.02246}, 2024.

\bibitem[Li et~al.(2024{\natexlab{a}})Li, Liu, Wu, Wang, Shen, Qu, Niu, Wang, Chen, and Li]{li2024aria}
Dongxu Li, Yudong Liu, Haoning Wu, Yue Wang, Zhiqi Shen, Bowen Qu, Xinyao Niu, Guoyin Wang, Bei Chen, and Junnan Li.
\newblock Aria: An open multimodal native mixture-of-experts model.
\newblock \emph{arXiv preprint arXiv:2410.05993}, 2024{\natexlab{a}}.

\bibitem[Li et~al.(2024{\natexlab{b}})Li, Zhang, Zhang, Zhang, Li, Li, Ma, and Li]{li2024llava}
Feng Li, Renrui Zhang, Hao Zhang, Yuanhan Zhang, Bo Li, Wei Li, Zejun Ma, and Chunyuan Li.
\newblock Llava-next-interleave: Tackling multi-image, video, and 3d in large multimodal models.
\newblock \emph{arXiv preprint arXiv:2407.07895}, 2024{\natexlab{b}}.

\bibitem[Li et~al.(2023{\natexlab{a}})Li, Li, Savarese, and Hoi]{li2023blip}
Junnan Li, Dongxu Li, Silvio Savarese, and Steven Hoi.
\newblock Blip-2: Bootstrapping language-image pre-training with frozen image encoders and large language models.
\newblock In \emph{International conference on machine learning}, pages 19730--19742. PMLR, 2023{\natexlab{a}}.

\bibitem[Li et~al.(2023{\natexlab{b}})Li, He, Wang, Li, Wang, Luo, Wang, Wang, and Qiao]{li2023videochat}
KunChang Li, Yinan He, Yi Wang, Yizhuo Li, Wenhai Wang, Ping Luo, Yali Wang, Limin Wang, and Yu Qiao.
\newblock Videochat: Chat-centric video understanding.
\newblock \emph{arXiv preprint arXiv:2305.06355}, 2023{\natexlab{b}}.

\bibitem[Li et~al.(2024{\natexlab{c}})Li, Wang, He, Li, Wang, Liu, Wang, Xu, Chen, Luo, et~al.]{li2024mvbench}
Kunchang Li, Yali Wang, Yinan He, Yizhuo Li, Yi Wang, Yi Liu, Zun Wang, Jilan Xu, Guo Chen, Ping Luo, et~al.
\newblock Mvbench: A comprehensive multi-modal video understanding benchmark.
\newblock In \emph{Proceedings of the IEEE/CVF Conference on Computer Vision and Pattern Recognition}, pages 22195--22206, 2024{\natexlab{c}}.

\bibitem[Li et~al.(2024{\natexlab{d}})Li, Chen, Hu, Wang, Shi, and Zhang]{li2024videovista}
Yunxin Li, Xinyu Chen, Baotian Hu, Longyue Wang, Haoyuan Shi, and Min Zhang.
\newblock Videovista: A versatile benchmark for video understanding and reasoning.
\newblock \emph{arXiv preprint arXiv:2406.11303}, 2024{\natexlab{d}}.

\bibitem[Li et~al.(2025)Li, Wang, and Jia]{li2025llama}
Yanwei Li, Chengyao Wang, and Jiaya Jia.
\newblock Llama-vid: An image is worth 2 tokens in large language models.
\newblock In \emph{European Conference on Computer Vision}, pages 323--340. Springer, 2025.

\bibitem[Lin et~al.(2023)Lin, Zhu, Ye, Ning, Jin, and Yuan]{lin2023video}
Bin Lin, Bin Zhu, Yang Ye, Munan Ning, Peng Jin, and Li Yuan.
\newblock Video-llava: Learning united visual representation by alignment before projection.
\newblock \emph{arXiv preprint arXiv:2311.10122}, 2023.

\bibitem[Liu et~al.(2024{\natexlab{a}})Liu, Li, Li, and Lee]{liu2024improved}
Haotian Liu, Chunyuan Li, Yuheng Li, and Yong~Jae Lee.
\newblock Improved baselines with visual instruction tuning.
\newblock In \emph{Proceedings of the IEEE/CVF Conference on Computer Vision and Pattern Recognition}, pages 26296--26306, 2024{\natexlab{a}}.

\bibitem[Liu et~al.(2024{\natexlab{b}})Liu, Wang, Ma, Wu, Ma, Wei, Jiao, Wu, and Hu]{liu2024kangaroo}
Jiajun Liu, Yibing Wang, Hanghang Ma, Xiaoping Wu, Xiaoqi Ma, Xiaoming Wei, Jianbin Jiao, Enhua Wu, and Jie Hu.
\newblock Kangaroo: A powerful video-language model supporting long-context video input.
\newblock \emph{arXiv preprint arXiv:2408.15542}, 2024{\natexlab{b}}.

\bibitem[Liu et~al.(2024{\natexlab{c}})Liu, Li, Tang, Ge, Shan, and St-llm]{liu2024large}
Ruyang Liu, Chen Li, Haoran Tang, Yixiao Ge, Ying Shan, and Ge~Li St-llm.
\newblock Large language models are effective temporal learners.
\newblock \emph{arXiv preprint arXiv:2404.00308}, 2024{\natexlab{c}}.

\bibitem[Liu et~al.(2024{\natexlab{d}})Liu, Li, Liu, Wang, Ren, Li, Chen, Sun, and Hou]{liu2024tempcompass}
Yuanxin Liu, Shicheng Li, Yi Liu, Yuxiang Wang, Shuhuai Ren, Lei Li, Sishuo Chen, Xu Sun, and Lu Hou.
\newblock Tempcompass: Do video llms really understand videos?
\newblock \emph{arXiv preprint arXiv:2403.00476}, 2024{\natexlab{d}}.

\bibitem[Luo et~al.(2023)Luo, Zhao, Yang, Dong, Li, Lu, Wang, Hu, Qiu, and Wei]{luo2023valley}
Ruipu Luo, Ziwang Zhao, Min Yang, Junwei Dong, Da Li, Pengcheng Lu, Tao Wang, Linmei Hu, Minghui Qiu, and Zhongyu Wei.
\newblock Valley: Video assistant with large language model enhanced ability.
\newblock \emph{arXiv preprint arXiv:2306.07207}, 2023.

\bibitem[Maaz et~al.(2023)Maaz, Rasheed, Khan, and Khan]{maaz2023video}
Muhammad Maaz, Hanoona Rasheed, Salman Khan, and Fahad~Shahbaz Khan.
\newblock Video-chatgpt: Towards detailed video understanding via large vision and language models.
\newblock \emph{arXiv preprint arXiv:2306.05424}, 2023.

\bibitem[Nan et~al.(2024)Nan, Xie, Zhou, Fan, Yang, Chen, Li, Yang, and Tai]{nan2024openvid}
Kepan Nan, Rui Xie, Penghao Zhou, Tiehan Fan, Zhenheng Yang, Zhijie Chen, Xiang Li, Jian Yang, and Ying Tai.
\newblock Openvid-1m: A large-scale high-quality dataset for text-to-video generation.
\newblock \emph{arXiv preprint arXiv:2407.02371}, 2024.

\bibitem[OpenAI(2023)]{OpenAI_ChatGPT_Website}
OpenAI.
\newblock Chatgpt.
\newblock \url{https://openai.com/index/chatgpt/}, 2023.

\bibitem[OpenAI(2024)]{OpenAI_GPT4o}
OpenAI.
\newblock Gpt-4o.
\newblock \url{https://openai.com/index/hello-gpt-4o/}, 2024.

\bibitem[Ordonez et~al.(2011)Ordonez, Kulkarni, and Berg]{ordonez2011im2text}
Vicente Ordonez, Girish Kulkarni, and Tamara Berg.
\newblock Im2text: Describing images using 1 million captioned photographs.
\newblock \emph{Advances in neural information processing systems}, 24, 2011.

\bibitem[Rajbhandari et~al.(2020)Rajbhandari, Rasley, Ruwase, and He]{rajbhandari2020zero}
Samyam Rajbhandari, Jeff Rasley, Olatunji Ruwase, and Yuxiong He.
\newblock Zero: Memory optimizations toward training trillion parameter models.
\newblock In \emph{SC20: International Conference for High Performance Computing, Networking, Storage and Analysis}, pages 1--16. IEEE, 2020.

\bibitem[Ren et~al.(2024)Ren, Yao, Li, Sun, and Hou]{ren2024timechat}
Shuhuai Ren, Linli Yao, Shicheng Li, Xu Sun, and Lu Hou.
\newblock Timechat: A time-sensitive multimodal large language model for long video understanding.
\newblock In \emph{Proceedings of the IEEE/CVF Conference on Computer Vision and Pattern Recognition}, pages 14313--14323, 2024.

\bibitem[Schuhmann et~al.(2022)Schuhmann, Beaumont, Vencu, Gordon, Wightman, Cherti, Coombes, Katta, Mullis, Wortsman, et~al.]{schuhmann2022laion}
Christoph Schuhmann, Romain Beaumont, Richard Vencu, Cade Gordon, Ross Wightman, Mehdi Cherti, Theo Coombes, Aarush Katta, Clayton Mullis, Mitchell Wortsman, et~al.
\newblock Laion-5b: An open large-scale dataset for training next generation image-text models.
\newblock \emph{Advances in Neural Information Processing Systems}, 35:\penalty0 25278--25294, 2022.

\bibitem[Sharma et~al.(2018)Sharma, Ding, Goodman, and Soricut]{sharma2018conceptual}
Piyush Sharma, Nan Ding, Sebastian Goodman, and Radu Soricut.
\newblock Conceptual captions: A cleaned, hypernymed, image alt-text dataset for automatic image captioning.
\newblock In \emph{Proceedings of the 56th Annual Meeting of the Association for Computational Linguistics (Volume 1: Long Papers)}, pages 2556--2565, 2018.

\bibitem[Shu et~al.(2024)Shu, Zhang, Liu, Qin, Zhou, Huang, and Zhao]{shu2024video}
Yan Shu, Peitian Zhang, Zheng Liu, Minghao Qin, Junjie Zhou, Tiejun Huang, and Bo Zhao.
\newblock Video-xl: Extra-long vision language model for hour-scale video understanding.
\newblock \emph{arXiv preprint arXiv:2409.14485}, 2024.

\bibitem[Song et~al.(2024)Song, Chai, Wang, Zhang, Zhou, Wu, Chi, Guo, Ye, Zhang, et~al.]{song2024moviechat}
Enxin Song, Wenhao Chai, Guanhong Wang, Yucheng Zhang, Haoyang Zhou, Feiyang Wu, Haozhe Chi, Xun Guo, Tian Ye, Yanting Zhang, et~al.
\newblock Moviechat: From dense token to sparse memory for long video understanding.
\newblock In \emph{Proceedings of the IEEE/CVF Conference on Computer Vision and Pattern Recognition}, pages 18221--18232, 2024.

\bibitem[Soomro(2012)]{soomro2012ucf101}
K Soomro.
\newblock Ucf101: A dataset of 101 human actions classes from videos in the wild.
\newblock \emph{arXiv preprint arXiv:1212.0402}, 2012.

\bibitem[Team et~al.(2023)Team, Anil, Borgeaud, Alayrac, Yu, Soricut, Schalkwyk, Dai, Hauth, Millican, et~al.]{team2023gemini}
Gemini Team, Rohan Anil, Sebastian Borgeaud, Jean-Baptiste Alayrac, Jiahui Yu, Radu Soricut, Johan Schalkwyk, Andrew~M Dai, Anja Hauth, Katie Millican, et~al.
\newblock Gemini: a family of highly capable multimodal models.
\newblock \emph{arXiv preprint arXiv:2312.11805}, 2023.

\bibitem[Wang et~al.(2024{\natexlab{a}})Wang, Shi, Tan, Qin, Wang, Zhang, Nambi, Ganu, and Wang]{wang2024multimodal}
Hengyi Wang, Haizhou Shi, Shiwei Tan, Weiyi Qin, Wenyuan Wang, Tunyu Zhang, Akshay Nambi, Tanuja Ganu, and Hao Wang.
\newblock Multimodal needle in a haystack: Benchmarking long-context capability of multimodal large language models.
\newblock \emph{arXiv preprint arXiv:2406.11230}, 2024{\natexlab{a}}.

\bibitem[Wang et~al.(2024{\natexlab{b}})Wang, Bai, Tan, Wang, Fan, Bai, Chen, Liu, Wang, Ge, et~al.]{wang2024qwen2}
Peng Wang, Shuai Bai, Sinan Tan, Shijie Wang, Zhihao Fan, Jinze Bai, Keqin Chen, Xuejing Liu, Jialin Wang, Wenbin Ge, et~al.
\newblock Qwen2-vl: Enhancing vision-language model's perception of the world at any resolution.
\newblock \emph{arXiv preprint arXiv:2409.12191}, 2024{\natexlab{b}}.

\bibitem[Wang et~al.(2024{\natexlab{c}})Wang, He, Hong, Cheng, Zhang, Qi, Huang, Xu, Dong, Ding, et~al.]{wang2024lvbench}
Weihan Wang, Zehai He, Wenyi Hong, Yean Cheng, Xiaohan Zhang, Ji Qi, Shiyu Huang, Bin Xu, Yuxiao Dong, Ming Ding, et~al.
\newblock Lvbench: An extreme long video understanding benchmark.
\newblock \emph{arXiv preprint arXiv:2406.08035}, 2024{\natexlab{c}}.

\bibitem[Wang et~al.(2024{\natexlab{d}})Wang, Zhang, Ren, Duan, Li, Liu, Hu, Chen, Zhang, Lu, et~al.]{wang2024needle}
Weiyun Wang, Shuibo Zhang, Yiming Ren, Yuchen Duan, Tiantong Li, Shuo Liu, Mengkang Hu, Zhe Chen, Kaipeng Zhang, Lewei Lu, et~al.
\newblock Needle in a multimodal haystack.
\newblock \emph{arXiv preprint arXiv:2406.07230}, 2024{\natexlab{d}}.

\bibitem[Wang et~al.(2024{\natexlab{e}})Wang, Song, Chen, Zhang, and Wang]{wang2024longllava}
Xidong Wang, Dingjie Song, Shunian Chen, Chen Zhang, and Benyou Wang.
\newblock Longllava: Scaling multi-modal llms to 1000 images efficiently via hybrid architecture.
\newblock \emph{arXiv preprint arXiv:2409.02889}, 2024{\natexlab{e}}.

\bibitem[Wang et~al.(2023)Wang, He, Li, Li, Yu, Ma, Li, Chen, Chen, Wang, et~al.]{wang2023internvid}
Yi Wang, Yinan He, Yizhuo Li, Kunchang Li, Jiashuo Yu, Xin Ma, Xinhao Li, Guo Chen, Xinyuan Chen, Yaohui Wang, et~al.
\newblock Internvid: A large-scale video-text dataset for multimodal understanding and generation.
\newblock \emph{arXiv preprint arXiv:2307.06942}, 2023.

\bibitem[Wu et~al.(2024)Wu, Li, Chen, and Li]{wu2024longvideobench}
Haoning Wu, Dongxu Li, Bei Chen, and Junnan Li.
\newblock Longvideobench: A benchmark for long-context interleaved video-language understanding.
\newblock \emph{arXiv preprint arXiv:2407.15754}, 2024.

\bibitem[Xiao et~al.(2021)Xiao, Shang, Yao, and Chua]{xiao2021next}
Junbin Xiao, Xindi Shang, Angela Yao, and Tat-Seng Chua.
\newblock Next-qa: Next phase of question-answering to explaining temporal actions.
\newblock In \emph{Proceedings of the IEEE/CVF conference on computer vision and pattern recognition}, pages 9777--9786, 2021.

\bibitem[Xu et~al.(2017)Xu, Zhao, Xiao, Wu, Zhang, He, and Zhuang]{xu2017video}
Dejing Xu, Zhou Zhao, Jun Xiao, Fei Wu, Hanwang Zhang, Xiangnan He, and Yueting Zhuang.
\newblock Video question answering via gradually refined attention over appearance and motion.
\newblock In \emph{Proceedings of the 25th ACM international conference on Multimedia}, pages 1645--1653, 2017.

\bibitem[Xu et~al.(2016)Xu, Mei, Yao, and Rui]{xu2016msr}
Jun Xu, Tao Mei, Ting Yao, and Yong Rui.
\newblock Msr-vtt: A large video description dataset for bridging video and language.
\newblock In \emph{Proceedings of the IEEE conference on computer vision and pattern recognition}, pages 5288--5296, 2016.

\bibitem[Xue et~al.(2024)Xue, Chen, Li, Hu, Zhu, Li, Fang, Tang, Yang, Liu, et~al.]{xue2024longvila}
Fuzhao Xue, Yukang Chen, Dacheng Li, Qinghao Hu, Ligeng Zhu, Xiuyu Li, Yunhao Fang, Haotian Tang, Shang Yang, Zhijian Liu, et~al.
\newblock Longvila: Scaling long-context visual language models for long videos.
\newblock \emph{arXiv preprint arXiv:2408.10188}, 2024.

\bibitem[Yang et~al.(2024)Yang, Yang, Hui, Zheng, Yu, Zhou, Li, Li, Liu, Huang, et~al.]{yang2024qwen2}
An Yang, Baosong Yang, Binyuan Hui, Bo Zheng, Bowen Yu, Chang Zhou, Chengpeng Li, Chengyuan Li, Dayiheng Liu, Fei Huang, et~al.
\newblock Qwen2 technical report.
\newblock \emph{arXiv preprint arXiv:2407.10671}, 2024.

\bibitem[Yao et~al.(2024)Yao, Yu, Zhang, Wang, Cui, Zhu, Cai, Li, Zhao, He, et~al.]{yao2024minicpm}
Yuan Yao, Tianyu Yu, Ao Zhang, Chongyi Wang, Junbo Cui, Hongji Zhu, Tianchi Cai, Haoyu Li, Weilin Zhao, Zhihui He, et~al.
\newblock Minicpm-v: A gpt-4v level mllm on your phone.
\newblock \emph{arXiv preprint arXiv:2408.01800}, 2024.

\bibitem[Yu et~al.(2019)Yu, Xu, Yu, Yu, Zhao, Zhuang, and Tao]{yu2019activitynet}
Zhou Yu, Dejing Xu, Jun Yu, Ting Yu, Zhou Zhao, Yueting Zhuang, and Dacheng Tao.
\newblock Activitynet-qa: A dataset for understanding complex web videos via question answering.
\newblock In \emph{Proceedings of the AAAI Conference on Artificial Intelligence}, pages 9127--9134, 2019.

\bibitem[Yun et~al.(2019)Yun, Han, Oh, Chun, Choe, and Yoo]{yun2019cutmix}
Sangdoo Yun, Dongyoon Han, Seong~Joon Oh, Sanghyuk Chun, Junsuk Choe, and Youngjoon Yoo.
\newblock Cutmix: Regularization strategy to train strong classifiers with localizable features.
\newblock In \emph{Proceedings of the IEEE/CVF international conference on computer vision}, pages 6023--6032, 2019.

\bibitem[Yun et~al.(2020)Yun, Oh, Heo, Han, and Kim]{yun2020videomix}
Sangdoo Yun, Seong~Joon Oh, Byeongho Heo, Dongyoon Han, and Jinhyung Kim.
\newblock Videomix: Rethinking data augmentation for video classification.
\newblock \emph{arXiv preprint arXiv:2012.03457}, 2020.

\bibitem[Zhang(2017)]{zhang2017mixup}
Hongyi Zhang.
\newblock mixup: Beyond empirical risk minimization.
\newblock \emph{arXiv preprint arXiv:1710.09412}, 2017.

\bibitem[Zhang et~al.(2023)Zhang, Li, and Bing]{zhang2023video}
Hang Zhang, Xin Li, and Lidong Bing.
\newblock Video-llama: An instruction-tuned audio-visual language model for video understanding.
\newblock \emph{arXiv preprint arXiv:2306.02858}, 2023.

\bibitem[Zhang et~al.(2024{\natexlab{a}})Zhang, Zhang, Li, Zeng, Yang, Zhang, Wang, Tan, Li, and Liu]{zhang2024long}
Peiyuan Zhang, Kaichen Zhang, Bo Li, Guangtao Zeng, Jingkang Yang, Yuanhan Zhang, Ziyue Wang, Haoran Tan, Chunyuan Li, and Ziwei Liu.
\newblock Long context transfer from language to vision.
\newblock \emph{arXiv preprint arXiv:2406.16852}, 2024{\natexlab{a}}.

\bibitem[Zhang et~al.(2024{\natexlab{b}})Zhang, Gui, Sun, Feng, Xu, Zhang, Fu, Li, Hauptmann, Bisk, et~al.]{zhang2024direct}
Ruohong Zhang, Liangke Gui, Zhiqing Sun, Yihao Feng, Keyang Xu, Yuanhan Zhang, Di Fu, Chunyuan Li, Alexander Hauptmann, Yonatan Bisk, et~al.
\newblock Direct preference optimization of video large multimodal models from language model reward.
\newblock \emph{arXiv preprint arXiv:2404.01258}, 2024{\natexlab{b}}.

\bibitem[Zhang et~al.(2024{\natexlab{c}})Zhang, Wu, Li, Li, Ma, Liu, and Li]{zhang2024video}
Yuanhan Zhang, Jinming Wu, Wei Li, Bo Li, Zejun Ma, Ziwei Liu, and Chunyuan Li.
\newblock Video instruction tuning with synthetic data.
\newblock \emph{arXiv preprint arXiv:2410.02713}, 2024{\natexlab{c}}.

\bibitem[Zhou et~al.(2024)Zhou, Shu, Zhao, Wu, Xiao, Yang, Xiong, Zhang, Huang, and Liu]{zhou2024mlvu}
Junjie Zhou, Yan Shu, Bo Zhao, Boya Wu, Shitao Xiao, Xi Yang, Yongping Xiong, Bo Zhang, Tiejun Huang, and Zheng Liu.
\newblock Mlvu: A comprehensive benchmark for multi-task long video understanding.
\newblock \emph{arXiv preprint arXiv:2406.04264}, 2024.

\bibitem[Zhou et~al.(2018)Zhou, Xu, and Corso]{zhou2018towards}
Luowei Zhou, Chenliang Xu, and Jason Corso.
\newblock Towards automatic learning of procedures from web instructional videos.
\newblock In \emph{Proceedings of the AAAI Conference on Artificial Intelligence}, 2018.

\end{thebibliography}
}

\clearpage
\setcounter{page}{1}
\maketitlesupplementary

\begin{figure*}[ht!]
  \centering
  \includegraphics[width=1.0\textwidth]{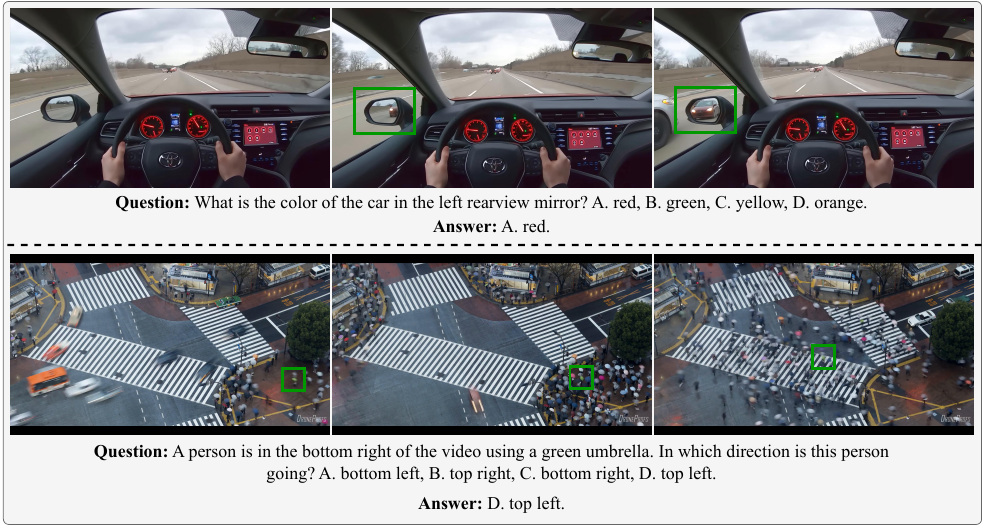}
  \vspace{-2em}
  \caption{Example questions from our \hrbench. Zoom in for better visualizations.}
  \label{fig:hrbench_example}
  \vspace{-5pt}
\end{figure*}

\section{\hrbench Details}
As detailed in Section~\ref{sec:hrbench}, our \hrbench consists of 200 questions covering 10 question types and 10 video types. That is, we collect two questions for each combination of video type and question type. The 10 video types are:
\begin{itemize}
    \item POV driving videos
    \item Egocentric sports videos
    \item Sportscast videos (broadcasting of sports events)
    \item Public event recordings
    \item Surveillance camera/CCTV footage
    \item Wildlife stock videos
    \item Aerial videos/Drone videos
    \item Factory and industrial stock videos
    \item Public transport videos
    \item Product review videos
\end{itemize}

For each question in the benchmark, we ensure the video duration falls between 3 to 10 seconds. This relatively short duration is chosen to maximize the likelihood of the frames relevant to the question getting sampled by the models. The final dataset has an average video duration of 5.4 seconds and an average resolution of 3048$\times$1699. Example questions and answers from our \hrbench are shown in Figure~\ref{fig:hrbench_example}.

\section{Model Training and Evaluation Details}
In this section, we provide additional details for training and benchmarking our selected baseline models.

\subsection{Baseline Models}
\textbf{VideoLLaVA} \cite{lin2023video} is a video LMM jointly pretrained on image and video data. It uses the pretrained Vicuna v1.5 model as its LLM backbone and LanguageBind as its image and video encoder. The model is pretrained on 558K image-text pairs from LAION-CC-SBU \cite{schuhmann2022laion, sharma2018conceptual, ordonez2011im2text} and 702K video-text pairs from Valley \cite{luo2023valley}. During the instruction-tuning stage, it incorporates 665K image-text pairs from LLaVA-1.5 \cite{liu2024improved} and 100K video-text pairs from Video-ChatGPT \cite{maaz2023video}.

\vspace{0.5em}
\noindent\textbf{Mantis-Idefics2} \cite{jiang2024mantis} is an LMM specialized in processing inputs with multiple interleaved images. It is initialized from Idefics2 \cite{laurenccon2024matters} and continually pretrained on Mantis-Instruct, a dataset comprising 721K interleaved image-text instruction-tuning examples. This dataset focuses on enhancing multi-image understanding across four dimensions: co-reference, comparison, reasoning, and temporal understanding. Mantis-Idefics2 achieves state-of-the-art performance on various multi-image benchmarks and excels on short video understanding benchmarks, such as MVBench \cite{li2024mvbench}.

\vspace{0.5em}
\noindent\textbf{LongVA} \cite{zhang2024long} is a long-context LMM designed for understanding long video content. It first performs continual pretraining using a Qwen2 \cite{yang2024qwen2} model to support up to 224K context length. Following this, it uses this modified Qwen2 model as the backbone for visual instruction tuning. LongVA is instruction-tuned on pure image data, using the same training data as LLaVA-1.6 \cite{liu2024improved}. It introduces the UniRes strategy, which divides an image into multiple grids and encodes each grid independently using the vision encoder. During inference, these grids are replaced by different frames from the input video, enabling effective processing of long video sequences.

\subsection{Additional Implementation Details}
For all three models, we conduct full-finetuning for one epoch using 8 Nvidia H800 GPUs. The total training time for $\sim$400K data is around one day. We use the Adam \cite{kingma2014adam} optimizer with a batch size of 128 during training. The learning rate is set to 5e-6 for VideoLLaVA and 1e-7 for LongVA and Mantis-Idefics2, with a cosine learning rate scheduler and a warm-up ratio of 0.03 applied to all models. We employ Flash-Attention 2 \cite{dao2023flashattention} and DeepSpeed ZeRO-3 \cite{rajbhandari2020zero} to accelerate training.

\subsection{Evaluation Benchmarks}
\noindent\textbf{Video-MME} \cite{fu2024video} is a comprehensive benchmark designed to evaluate the video analysis capabilities of LMMs. It includes 900 videos and 2700 questions across six visual domains. The questions are categorized based on video durations into short, medium, and long video questions, with median durations of 26s, 164.7s, and 890.7s, respectively. The median duration values for short, medium and long video questions are 26s, 164.7s, and 890.7s, respectively. Video-MME supports two evaluation formats: (1) the ``w/ subtitle'' format, which includes both the video subtitles and questions as text inputs, and (2) the ``w/o subtitle'' format, which uses only the raw video and questions as inputs. In the main paper, we focus on the ``w/o subtitle'' format to emphasize improving the long video understanding capabilities of video LMMs through video augmentation, rather than relying on additional subtitle information. For completeness, we provide results for the ``w/ subtitle'' format in Section~\ref{sec:additional_experiment}.

\vspace{0.5em}
\noindent\textbf{MLVU} \cite{zhou2024mlvu} is a long video understanding benchmark encompassing diverse tasks and video genres. It features two types of questions: multiple-choice questions and freeform generation questions. The benchmark evaluates LMMs across three dimensions: holistic video understanding, requiring global information from the entire video; single-detail video understanding, focused on short and salient moments within the video; and multi-detail video understanding, involving connections across multiple short clips in the video. In this paper, we report the accuracy scores for the multiple-choice questions from the development set of MLVU. In the paper, we report the accuracy scores for the multiple-choice questions from the dev set of MLVU.

\vspace{0.5em}
\noindent\textbf{LVBench} \cite{wang2024lvbench} evaluates the comprehension capabilities of video LMMs for extremely long videos. It consists of 1549 QA pairs, with an average video duration of 4101 seconds. The benchmark assesses video LMMs across six core aspects: temporal grounding, video summarization, video reasoning, entity recognition, event understanding, and key information retrieval. We use the full test set for evaluation.

\vspace{0.5em}
\noindent\textbf{LongVideoBench} \cite{wu2024longvideobench} is a question-answering benchmark featuring interleaved long video-text input. The dataset contains 3763 videos and 6678 human-annotated multiple-choice questions spanning 17 fine-grained categories. LongVideoBench supports two evaluation formats: the standard input format, where video tokens are processed first followed by question descriptions, and an interleaved video-text format, where subtitles are inserted between video frames. Although Mantis-Idefics2 supports interleaved image-text input, as our \dataset does not include training examples in such format, we still evaluate Mantis-Idefics2 and the finetuned \model-Mantis using the standard format. We report the results of the validation split.

\vspace{0.5em}
\noindent\textbf{MVBench and NExT-QA} \cite{li2024mvbench, xiao2021next} are short video understanding benchmarks, focusing on videos under one minute in duration. MVBench includes 4,000 multiple-choice questions derived from 3,641 video clips, with an average video duration of 16 seconds. NExT-QA comprises 8,564 questions (both multiple-choice and open-ended) sourced from 1,000 videos, averaging 40 seconds in length. In our experiments, we evaluate the models on the full MVBench dataset and the MCQ split of NExT-QA.

\begin{table*}[]
\centering
\small
\caption{Comparison between the baseline VideoLLaVA model, VideoLLaVA finetuned on \dataset and VideoLLaVA finetuned on \dataset + 300K VideoChat2-IT data (\model-VideoLLaVA in the main paper) on long video understanding benchmarks. ``SFT'' indicates supervised finetuning.}
\setlength\tabcolsep{4 pt}
\begin{tabular}{@{}l|ccccccc@{}}
\toprule
                                                                    & \multicolumn{7}{c}{Long Video Understanding}                                                                                                                                                                                                                                                                                                  \\ \cmidrule(l){2-8} 
                                                                    & \multicolumn{4}{c|}{Video-MME w/o subtitles}                                                                                                                                   & \multicolumn{1}{c|}{MLVU}                                 & \multicolumn{1}{c|}{LVBench}                              & LongVideoBench                       \\ \cmidrule(l){2-8} 
\multirow{-3}{*}{Models}                                            & avg                                  & short                                & medium                               & \multicolumn{1}{c|}{long}                                 & \multicolumn{1}{c|}{m-avg}                                & \multicolumn{1}{c|}{test}                                 & val                                  \\ \midrule
VideoLLaVA                                                          & 39.9                                 & 45.3                                 & 38.0                                 & \multicolumn{1}{c|}{36.2}                                 & \multicolumn{1}{c|}{45.0}                                 & \multicolumn{1}{c|}{29.3}                                 & 39.1                                 \\ \midrule
VideoLLaVA (SFT on \dataset)                         & 43.6                                 & 47.3                                 & 43.8                                 & \multicolumn{1}{c|}{39.8}                                 & \multicolumn{1}{c|}{48.7}                                 & \multicolumn{1}{c|}{32.6}                                 & 41.0                                 \\
\rowcolor{LightCyan}
$\Delta$ - VideoLLaVA                                  & {\color[HTML]{009901} \textbf{+3.7}} & {\color[HTML]{009901} \textbf{+2.0}} & {\color[HTML]{009901} \textbf{+5.8}} & \multicolumn{1}{c|}{{\color[HTML]{009901} \textbf{+3.6}}} & \multicolumn{1}{c|}{{\color[HTML]{009901} \textbf{+3.7}}} & \multicolumn{1}{c|}{{\color[HTML]{009901} \textbf{+3.3}}} & {\color[HTML]{009901} \textbf{+1.9}} \\ \midrule
VideoLLaVA (SFT on \dataset + 300K VideoChat2-IT)    & 43.7                                 & 48.2                                 & 43.9                                 & \multicolumn{1}{c|}{38.9}                                 & \multicolumn{1}{c|}{49.5}                                 & \multicolumn{1}{c|}{33.8}                                 & 42.3                                 \\
\rowcolor{LightCyan}
$\Delta$ - VideoLLaVA (SFT on \dataset) & {\color[HTML]{009901} \textbf{+0.1}} & {\color[HTML]{009901} \textbf{+0.9}} & {\color[HTML]{009901} \textbf{+0.1}} & \multicolumn{1}{c|}{{\color[HTML]{FE0000} \textbf{-0.9}}} & \multicolumn{1}{c|}{{\color[HTML]{009901} \textbf{+0.8}}} & \multicolumn{1}{c|}{{\color[HTML]{009901} \textbf{+1.2}}} & {\color[HTML]{009901} \textbf{+1.3}} \\ \bottomrule
\end{tabular}
\label{tab:videollava_ablation_long}
\end{table*}
\begin{table*}[]
\centering
\small
\caption{Comparison between the baseline VideoLLaVA model, VideoLLaVA finetuned on \dataset and VideoLLaVA finetuned on \dataset + 300K VideoChat2-IT data (\model-VideoLLaVA in the main paper) on \hrbench. ``SFT'' indicates supervised finetuning.}
\setlength\tabcolsep{7 pt}
\begin{tabular}{@{}l|ccc@{}}
\toprule
                                                                    & \multicolumn{3}{c}{High-Resolution Video Understanding}                                                              \\ \cmidrule(l){2-4} 
                                                                    & \multicolumn{3}{c}{\hrbench}                                                                                     \\ \cmidrule(l){2-4} 
\multirow{-3}{*}{Models}                                            & avg                                   & object                               & action                                \\ \midrule
VideoLLaVA                                                          & 32.5                                  & 36.0                                 & 27.9                                  \\ \midrule
VideoLLaVA (SFT on \dataset)                         & 44.0                                  & 42.1                                 & 46.5                                  \\
\rowcolor{LightCyan}
$\Delta$ - VideoLLaVA                                  & {\color[HTML]{009901} \textbf{+11.5}} & {\color[HTML]{009901} \textbf{+6.1}} & {\color[HTML]{009901} \textbf{+18.6}} \\ \midrule
VideoLLaVA (SFT on \dataset + 300K VideoChat2-IT)    & 47.5                                  & 50                                   & 44.2                                  \\
\rowcolor{LightCyan}
$\Delta$ - VideoLLaVA (SFT on \dataset) & {\color[HTML]{009901} \textbf{+3.5}}  & {\color[HTML]{009901} \textbf{+7.9}} & {\color[HTML]{FE0000} \textbf{-2.3}}  \\ \bottomrule
\end{tabular}
\label{tab:videollava_ablation_high}
\end{table*}

\vspace{0.5em}
\noindent\textbf{MSVD-QA, MSRVTT-QA, TGIF-QA and ActivityNet-QA} \cite{xu2017video, jang2017tgif, yu2019activitynet} are open-ended QA benchmarks designed to evaluate the response generation capabilities of video LMMs. These benchmarks consist of short videos and assess the ability of video LMMs to produce simple, coherent answers. For all four benchmarks, we follow Video-ChatGPT \cite{maaz2023video} and use GPT-3.5-Turbo to evaluate the accuracy and quality of the responses. Specifically, GPT is prompted with the ground truth answer and the model's response to determine if the answer is correct (yes/no) and to assign a quality score between 1 and 5. Following Video-ChatGPT, we evaluate the models on the validation sets of MSVD-QA, MSRVTT-QA and ActivityNet-QA, and use the FrameQA split from TGIF-QA's test set for evaluation. Since GPT-3.5-Turbo's API version has changed and the older API versions are no longer accessible, we are unable to reproduce the results for some baseline models. In the paper, we report all scores based on our evaluation script.

\section{Additional Experimental Results}
\label{sec:additional_experiment}

\subsection{Training Data Ablations for VideoLLaVA}
As mentioned in Section~\ref{sec:eval_setup}, unlike Mantis-Idefics2 and LongVA, we fine-tune VideoLLaVA using a combination of our \dataset and 300K short video samples from VideoChat2-IT to preserve its short video understanding capabilities. In this section, we examine how this additional training data impacts the model's performance on long and high-resolution video understanding tasks after finetuning. To assess this, we finetune another VideoLLaVA model exclusively on our \dataset and compare the results against the combined training approach in Table~\ref{tab:videollava_ablation_long} and Table~\ref{tab:videollava_ablation_high}.

As shown in Table~\ref{tab:videollava_ablation_long}, finetuning VideoLLaVA exclusively on our \dataset results in consistent improvements across all long video understanding benchmarks. On the other hand, incorporating an additional 300K short video samples does not yield further significant gains in long video understanding. Notably, the Video-MME results indicate that adding short video data slightly detracts from the model's performance on long videos, underscoring the importance of our dataset for enhancing long video understanding capabilities.

For high-resolution video understanding, according to Table~\ref{tab:videollava_ablation_high}, finetuning VideoLLaVA on our data leads to a significant improvement (+11.5\%) on \hrbench. While adding additional short video data further enhances model performance, the improvement is less substantial. These findings suggest that our dataset remains the primary driver of performance gains in high-resolution video understanding. Moreover, incorporating VideoChat2-IT training data leads to a decline in performance on action-related questions, highlighting the superior effectiveness of our dataset for tasks requiring temporal understanding.

\subsection{Video-MME w/ Subtitles Results}
\begin{table}[]
\centering
\small
\caption{Comparison between \model-finetuned models and baseline models on Video-MME w/ subtitle benchmark.}
\setlength\tabcolsep{7 pt}
\begin{tabular}{@{}l|cccc@{}}
\toprule
                                        & \multicolumn{4}{c}{Video-MME w/ subtitles}                                                                                                                \\ \cmidrule(l){2-5} 
\multirow{-2}{*}{Models}                & avg                                  & short                                & medium                               & long                                 \\ \midrule
VideoLLaVA                              & 41.6                                 & 46.1                                 & 40.7                                 & 38.1                                 \\
\model-VideoLLaVA        & 45.1                                 & 50.2                                 & 45.7                                 & 39.3                                 \\
\rowcolor{LightCyan}
$\Delta$ - VideoLLaVA      & {\color[HTML]{009901} \textbf{+3.5}} & {\color[HTML]{009901} \textbf{+4.1}} & {\color[HTML]{009901} \textbf{+5.0}} & {\color[HTML]{009901} \textbf{+1.2}} \\ \midrule
Mantis-Idefics2                         & 49.0                                 & 60.4                                 & 46.1                                 & 40.3                                 \\
\model-Mantis            & 50.9                                 & 61.8                                 & 48.6                                 & 42.3                                 \\
\rowcolor{LightCyan}
$\Delta$ - Mantis-Idefics2 & {\color[HTML]{009901} \textbf{+1.9}} & {\color[HTML]{009901} \textbf{+1.4}} & {\color[HTML]{009901} \textbf{+2.5}} & {\color[HTML]{009901} \textbf{+2.0}} \\ \midrule
LongVA                                  & 54.3                                 & 61.6                                 & 53.6                                 & 47.6                                 \\
\model-LongVA            & 59.3                                 & 70.0                                 & 57.6                                 & 50.3                                 \\
\rowcolor{LightCyan}
$\Delta$ - LongVA          & {\color[HTML]{009901} \textbf{+5.0}} & {\color[HTML]{009901} \textbf{+8.4}} & {\color[HTML]{009901} \textbf{+4.0}} & {\color[HTML]{009901} \textbf{+2.7}} \\ \bottomrule
\end{tabular}
\label{tab:videomme_w_subtitle}
\end{table}
We show the results for Video-MME w/ subtitles in Table~\ref{tab:videomme_w_subtitle}. In this evaluation setting, the video's subtitles are provided as part of the question input to the model. The results indicate that both baseline models and our \model-finetuned models can be further enhanced by providing extra subtitle information. Similar to Video-MME w/o subtitles results, our \model-finetuned models consistently achieve better performances compared to the baseline models. This shows that our synthetic data provides consistent and model-agnostic enhancements to the long video understanding capability of video LMMs.

\section{Limitations}
Our method exhibits a few limitations. First, since we generate instruction data based on video captions, and most public video-caption datasets contain simple captions for video clips, our synthesized data often contain short responses, leading to a shorter response from the finetuned models. This issue could be addressed by recaptioning the raw video data using high-capacity video captioning models. Second, while our synthesized augmented video data have been shown to enhance long and high-resolution video understanding, the current video augmentation paradigm does not fully align with real-world video distributions. Addressing this limitation would require more advanced video combination and blending techniques, such as leveraging segmentation maps to isolate specific regions from one video and seamlessly integrating them into another to create more natural and realistic augmented video samples.

\section{Instruction Synthesis Prompt Templates}
In this section, we list the Gemini prompts we used to synthesize instruction data below.
\NewTColorBox{Long_Caption}{ s O{!htbp} }{%
  floatplacement={#2},
  IfBooleanTF={#1}{float*,width=\textwidth}{float},
  colframe=orange!50!black,colback=orange!10!white,title=Long Video Caption Generation Prompt,
  }

\NewTColorBox{Event_QA_Generation}{ s O{!htbp} }{%
  floatplacement={#2},
  IfBooleanTF={#1}{float*,width=\textwidth}{float},
  colframe=green!50!black,colback=green!10!white,title=Event Relationship QA Generation Prompt,
  }

\NewTColorBox{QA_Generation}{ s O{!htbp} }{%
  floatplacement={#2},
  IfBooleanTF={#1}{float*,width=\textwidth}{float},
  colframe=yellow!50!black,colback=yellow!10!white,title=Freeform QA Generation Prompt,
  }

\NewTColorBox{MCQ_Generation}{ s O{!htbp} }{%
  floatplacement={#2},
  IfBooleanTF={#1}{float*,width=\textwidth}{float},
  colframe=cyan!50!black,colback=cyan!10!white,title=MCQ Generation Prompt,
  }

\begin{QA_Generation}[!ht]
 
\textbf{User:}

Given a short paragraph of caption describing a video clip, can you try to extract relevant information from the caption and come up with a question-answer pair that could possibly reflect the facts of some local and fine-grained scenes in the video?\\
The caption of the video is as follows:\\
$<$Video Caption$>$\\
\\
Please try not to come up with questions that you cannot answer. Please also note that the caption will not be presented in the actual training data. Return only the question and the answer. Format your output as:\\
\#\#\#Question\#\#\#\\
$<$your question$>$\\
\#\#\#Answer\#\#\#\\
$<$your answer$>$
\\
\\
\textbf{Assistant:}

$<$Synthesized Freeform QA pairs$>$
\end{QA_Generation}

\begin{MCQ_Generation}[!ht]
 
\textbf{User:}

Given the following Question-Answer pair, turn this short answer question into a multiple-choice question by synthesizing three additional incorrect options. Assume the correct option is $<$Random Option between A to D$>$. \\

Question:

$<$Question$>$

Answer:

$<$Freeform Answer$>$
\\
\\
Your output should be in the format of a python list:

[\\
``A. $<$answer1$>$'',\\
``B. $<$answer2$>$'',\\
``C. $<$answer3$>$'',\\
``D. $<$answer4$>$''\\
]\\
\\
\textbf{Assistant:}

$<$Synthesized MCQ pairs$>$
\end{MCQ_Generation}

\begin{Event_QA_Generation}[!ht]
 
\textbf{User:}

Given multiple short captions, each representing a short chunk of video in a longer video, generate a question-answer pair related to the order of the events in the video. Note that because the short captions are from the same video, you can combine entities with slightly different descriptions in different captions, as they most likely represent the same thing. Format the output using the following format:\\
\#\#\#Question\#\#\#\\
$<$Your question$>$\\
\#\#\#Answer\#\#\#\\
$<$Your answer$>$\\
\\
For example, given captions like:\\
Caption 1: A squirrel is sitting on a tree branch in a forest, surrounded by pine trees and blue sky.\\
Caption 2: A cartoon squirrel is holding an egg in a tree.\\
Caption 3: A cartoon squirrel is standing next to an egg.\\
Your output can be:\\
\#\#\#Question\#\#\#\\
What happens after the squirrel sits on a tree branch?\\
\#\#\#Answer\#\#\#\\
The squirrel holds an egg.\\
\\
Try to be creative with your question and answer.\\
\\
The short captions (in chronological order) are listed below:\\
Caption 1. $<$Caption 1$>$\\
Caption 2. $<$Caption 2$>$\\
...\\
Caption N. $<$Caption N$>$\\
\\
\\
\textbf{Assistant:}

$<$Synthesized Event Relationship QA pairs$>$
\end{Event_QA_Generation}

\begin{Long_Caption}[!ht]
 
\textbf{User:}

Given multiple short captions, each representing a short chunk of video in a longer video, create a detailed caption by combining the short captions such that the detailed caption describes the whole video. Note that because the short captions are from the same video, you can combine entities with slightly different descriptions in different captions, as they most likely represent the same thing. Return only the caption.\\
\\
The short captions (in chronological order) are listed below:\\
Caption 1. $<$Caption 1$>$\\
Caption 2. $<$Caption 2$>$\\
...\\
Caption N. $<$Caption N$>$\\
\\
\textbf{Assistant:}

$<$Synthesized Long Video Caption$>$
\end{Long_Caption}

\end{document}